%% file: MFKAN.tex
\documentclass[review]{elsarticle}

\usepackage{hyperref}
\usepackage{longtable}
\usepackage{amsmath,amsthm,amssymb,epsfig,bm,amsmath}
 \usepackage{graphicx}
\usepackage{float}
\usepackage{amsmath}
\usepackage{mhchem}
\usepackage{xparse}
\usepackage{adjustbox}
\usepackage{MnSymbol}
\usepackage{mathdots}
\usepackage{makecell}
\usepackage{listings}
\usepackage{footnote}
\usepackage[flushleft]{threeparttable}
\usepackage{multirow}
\usepackage{relsize}
\usepackage{lettrine}
\usepackage{courier}
\usepackage{color} 
\definecolor{mygreen}{RGB}{28,172,0} 
\definecolor{mylilas}{RGB}{170,55,241}
\usepackage{amsmath,amsthm,amssymb,amsfonts} 
\usepackage{graphicx} 
\usepackage[margin=1in]{geometry} 
\usepackage[yyyymmdd]{datetime}
\usepackage{algorithm}
\usepackage[noend]{algpseudocode}
\usepackage{lipsum}
\usepackage{setspace}
\usepackage{siunitx}
\usepackage{pdfpages}
\usepackage{tikz}
\usetikzlibrary{positioning}
\usepackage{booktabs,caption}
\usepackage[]{hyperref}
\usepackage{soul}
\usepackage{xcolor}
\usepackage{subcaption}

\hypersetup{
 colorlinks  = true, 
 urlcolor   = blue, 
 linkcolor  = black, 
 citecolor  = blue 
}
\usepackage[numbers]{natbib}
\usepackage{color}
\definecolor{codegreen}{rgb}{0,0.6,0}
\definecolor{codegray}{rgb}{0.5,0.5,0.5}
\definecolor{codepurple}{rgb}{0.58,0,0.82}
\definecolor{backcolour}{rgb}{0.95,0.95,0.92}
\lstdefinestyle{mystyle}{
  backgroundcolor=\color{backcolour},
  commentstyle=\color{codegreen},
  keywordstyle=\color{magenta},
  numberstyle=\tiny\color{codegray},
  stringstyle=\color{codepurple},
  basicstyle=\footnotesize,
  breakatwhitespace=false,
  breaklines=true,
  captionpos=b,
  keepspaces=true,
  numbers=left,
  numbersep=5pt,
  showspaces=false,
  showstringspaces=false,
  showtabs=false,
  tabsize=2,
  escapeinside={<@}{@>},
}

\usepackage{booktabs} 
\usepackage{siunitx} 
\usepackage{comment}
\usepackage{mathtools} 
\usepackage{gensymb} 
\usepackage{mhchem} 
\usepackage{fixltx2e} 
\usepackage{bbm}

\usepackage{multirow}

\usepackage{fancyhdr} 
\theoremstyle{definition}

\theoremstyle{definition}

\theoremstyle{remark}

\usepackage{soul} 
\soulregister\cite7
\soulregister\ref7
\soulregister\pageref7
\usepackage{bm}

\usepackage{framed} 

\usepackage{multicol} 

\usepackage{nomencl} 
\usepackage{tikz}
\makenomenclature
\usepackage[resetlabels,labeled]{multibib}

\renewcommand*\nompreamble{\begin{multicols}{2}}
\renewcommand*\nompostamble{\end{multicols}}

\usepackage{amssymb}
\usepackage{pifont}
\definecolor{light-gray}{gray}{0.95}

\usepackage[left, pagewise]{lineno}

\journal{Elsevier}

\begin{document}


\begin{frontmatter}


\title{\large Interpretable Multifidelity Learning with Kolmogorov-Arnold Networks for Spatiotemporal Nuclear Reactor Flow Modeling}

\author{Meredith Eaheart$^{a*}$, Majdi I. Radaideh$^{a,b*}$}

\cortext[mycorrespondingauthor]{Corresponding Author: M. Eaheart (eaheart@umich.edu), M. I. Radaideh (radaideh@umich.edu)}

\address{$^{a}$Department of Nuclear Engineering and Radiological Science, University of Michigan, Ann Arbor, Michigan 48109}
\address{$^{b}$Department of Computer Science and Engineering, University of Michigan, Ann Arbor, Michigan 48109}

\begin{abstract}
\small

High-fidelity computational fluid dynamics (CFD) simulations of nuclear reactor systems are computationally expensive, motivating machine learning methods that reduce cost while preserving accuracy. Multifidelity surrogate modeling addresses this by combining inexpensive low-fidelity data with limited high-fidelity data. This work investigates whether the high-fidelity data required to predict the spatiotemporal mass flow rate across all 193 fuel assemblies of a pressurized water reactor (PWR) core can be reduced, comparing black-box multifidelity feedforward neural networks (MFNNs) with interpretable multifidelity Kolmogorov-Arnold Networks (MFKANs). Cross-fidelity correlation across the core is weak, making this a challenging multifidelity problem, and neither architecture predicts the transient particularly well in absolute terms. MFKAN nonetheless performs comparably to MFNN for one-time-step-forward prediction while also providing closed-form symbolic equations unavailable with MFNN. To address the scarcity of CFD transients, with only one available per fidelity level, TabDDPM-generated synthetic data was used to augment cross-fidelity forecasting. KAN and FNN achieve comparable self-evaluation accuracy (Pearson correlation of roughly 0.73--0.81), but both perform poorly at true cross-fidelity forecasting (correlations of 0.03--0.06). Assembly-level analysis shows that each assembly's forecast correlation tracks its real cross-fidelity correlation for both architectures ($r=0.840$--$0.896$ for KAN, $r=0.804$--$0.903$ for FNN), confirming that forecasting accuracy is fundamentally limited by how correlated the fidelity levels already are. Overall, KAN-based multifidelity modeling achieves accuracy consistently comparable to MFNN across all settings, while additionally providing symbolic equations that match or exceed the spline representation's accuracy.

\end{abstract}


\end{frontmatter}

\setstretch{1.3}

\section{Introduction}
\label{sec:intro}

Nuclear reactors experience complex physical phenomena and require detailed
simulations to capture this complexity, but that comes at a cost: high-fidelity computational fluid dynamics (CFD) simulations
are accurate but expensive to run~\citep{price2023thermal}, while lower-fidelity simulations are
cheaper but less accurate. This tradeoff motivates machine learning methods
that can preserve accuracy while cutting the computational cost
\citep{burnett2026high}. One technique that has been used to save computational cost is surrogate modeling and more specifically multifidelity surrogate modeling, which combines a small
amount of expensive high-fidelity data with a larger amount of cheap low-fidelity data.

The problem considered in this work combines several difficulties that are rarely encountered together. The quantity of interest is the full spatiotemporal mass flow rate field at the core inlet of a pressurized water reactor (PWR), a high-dimensional, turbulence-driven transient, and the available CFD fidelity levels are only weakly correlated with one another. The CFD simulations used for multifidelity surrogate modeling in this work are also limited to a single transient at each fidelity level, restricting the amount and diversity of data available for training, and generating additional high-fidelity CFD transients is computationally expensive. At the same time, safety-relevant applications favor surrogate models whose predictions can be inspected rather than taken on trust~\citep{rudin2019stop}. Section~\ref{sec:related} reviews how existing work addresses each of these difficulties and identifies the gaps that remain.

This work develops and systematically evaluates multifidelity feedforward neural networks (MFNNs) and multifidelity Kolmogorov-Arnold Networks (MFKANs) for predicting the full spatiotemporal mass flow rate field in a large-scale PWR core. Three CFD mesh-based fidelity levels derived from the same underlying reactor model \citep{burnett2026high} are used to investigate multifidelity learning under weak cross-fidelity correlation and severely limited transient data. Beyond benchmarking established multifidelity neural-network approaches, this work introduces new KAN-based architectures designed to combine multifidelity prediction with symbolic interpretability. The main contributions are:

\begin{itemize}
    \item Full-core spatiotemporal multifidelity prediction: MFNN and MFKAN methods are applied, for the first time, to full-core spatiotemporal mass flow rate prediction across all 193 PWR fuel assemblies, extending multifidelity surrogate modeling from scalar or low-dimensional quantities to a high-dimensional, time-dependent reactor field.

    \item New fidelity-conditioned KAN architecture: A new multifidelity architecture, \textit{MFKAN-Flag}, is introduced in which the fidelity level is explicitly encoded as an input to a unified KAN. MFKAN-Flag is benchmarked against MFKAN-Delta and conventional MFNN approaches to determine whether KAN-based multifidelity models can achieve comparable predictive accuracy while additionally yielding interpretable symbolic representations unavailable from standard black-box neural networks.

    \item Extension to three fidelity levels: \textit{MFKAN-Flag3f} extends the proposed fidelity-conditioned architecture from two to three CFD fidelity levels, demonstrating how a single KAN framework can simultaneously learn information across multiple mesh resolutions rather than being restricted to a conventional low-/high-fidelity pair.

    \item Generative augmentation for data-scarce multifidelity CFD:
    TabDDPM-based synthetic data generation is integrated with the multifidelity framework to address the extreme scarcity of CFD transient data, with only one transient available at each fidelity level. The augmented datasets are used for cross-fidelity transient forecasting, providing a pathway to exploit generative models when additional high-fidelity CFD simulations are prohibitively expensive.
\end{itemize}

These contributions establish and evaluate an interpretable KAN-based framework for high-dimensional multifidelity reactor prediction, while addressing two central challenges of practical CFD surrogate modeling: weak correlations between fidelity levels and severe scarcity of high-fidelity transient data. The remainder of this paper is organized as follows. Section~\ref{sec:related} reviews related work and identifies the gaps addressed here. Section~\ref{sec:methods} describes the dataset and the machine learning and multifidelity methods considered in this work. Section~\ref{sec:results} presents the results of the corresponding experiments, while Section~\ref{sec:discussion} discusses the key findings, study limitations, and directions for future work, and Section~\ref{sec:conclusions} summarizes the main conclusions.

\section{Related Work}
\label{sec:related}

Surrogate modeling is a widely used machine learning technique for reducing
the computational cost of expensive physics-based simulations~\citep{radaideh2019combining,radaideh2020surrogate,myers2025pymaise}, and it has been
applied across a wide range of problems in nuclear engineering, including
model calibration \citep{radaideh2022model}, design under uncertainty constraints \citep{erdem2025surrogate}, and design and operation optimization \citep{sobes2021ai}. Optimization studies in particular have used evolutionary, swarm, and reinforcement-learning-guided algorithms for tasks such as nuclear fuel design \citep{radaideh2022pesa}, microreactor reactivity control system operation \citep{price2022multiobjective}, and general neuroevolution-based engineering design \citep{radaideh2021neorl}. Because these algorithms typically require thousands of model evaluations, fast surrogates are often what makes them computationally feasible. Multifidelity
surrogate modeling methods extend this idea further by combining a small
amount of expensive high-fidelity data with a larger amount of cheap low
fidelity data, an approach with a long history in engineering more broadly
\citep{fernandez2016review,peherstorfer2018survey}.
These multifidelity surrogate models are most commonly built as either
multifidelity neural networks (MFNN) \citep{meng2020composite, guo2022multi, yi2024practical} or multifidelity Gaussian processes \citep{kennedy2000predicting,perdikaris2017nonlinear}.
Our own prior work directly compared these
two architectures for a nuclear thermal-hydraulics application \citep{eaheart2026multifidelity}.

Multifidelity surrogate modeling has also been applied extensively outside
nuclear engineering, particularly in fluid mechanics, where it has supported
tasks such as aerodynamic shape optimization \citep{pena2012surrogate}, turbomachinery flow
prediction \citep{li2023multi}, and CFD-based thermal design optimization \citep{ghosh2024parametric}. Within nuclear engineering specifically, multifidelity methods have been used for thermal hydraulic analysis \citep{hanna2020machine,eaheart2026multifidelity} and reactor core simulations \citep{oktavian2023preliminary}.

Multifidelity modeling is not beneficial in every setting, however. Systematic studies have shown that its advantage over single-fidelity surrogates depends strongly on how well the fidelity levels are correlated, and can disappear when the low-fidelity model is a poor predictor of the high-fidelity one \citep{toal2015some,fernandezgodino2019issues}. Nuclear multifidelity applications to date have mostly targeted scalar or low-dimensional quantities of interest, and little is known about how these methods behave when cross-fidelity correlation is weak.

The coolant flow distribution at the inlet of a PWR core is shaped by mixing in the downcomer and lower plenum, and its prediction has long relied on CFD validated against dedicated experiments \citep{rohde2007fluid}. Resolving the turbulence-driven fluctuations of this flow across a full core at fine temporal resolution is particularly expensive \citep{burnett2026high}. Machine learning has become a common route to reducing such costs in fluid mechanics \citep{brunton2020machine}, typically by compressing the flow field into a small number of modes and learning how those modes evolve in time. These reduced-order models follow two broad designs: static feedforward networks that map the current modal state to the next one \citep{lui2019construction}, and recurrent architectures, such as long short-term memory (LSTM) networks, that carry information across longer histories \citep{hasegawa2020cnn,radaideh2020neural,radaideh2022time}.

In nuclear engineering, deep neural networks and reduced-order models have been used to predict high-dimensional, core-wide quantities such as power distributions, neutron flux, and three-dimensional temperature fields \citep{saleem2020application,gong2022data,price2024simplified}. For the same PWR lower-plenum model used in this work, machine learning has also been applied to flow-field reconstruction and one-step-ahead forecasting \citep{burnett2026high}. These high-dimensional studies, however, rely on a single source of training data. To our knowledge, multifidelity learning has not yet been applied to full-core spatiotemporal fields, and the trade-off between static and recurrent architectures has not been considered in settings where the resulting model must also be interpretable.

Despite these successes, the multifidelity methods described above are
typically built on classical feedforward neural networks, which offer
little insight into why a given prediction was made \citep{panczyk2025opening},
or Gaussian processes, which provide useful uncertainty quantification
but not the same closed-form interpretability offered by symbolic
regression. For safety-relevant applications, models that are interpretable by construction are generally preferable to post hoc explanations of black-box models \citep{rudin2019stop}. Symbolic approaches pursue this goal by expressing the learned model as an explicit equation, ranging from polynomial networks built with the group method of data handling \citep{radaideh2020analyzing} to sparse identification of governing equations \citep{brunton2016discovering} and neural-network-guided symbolic regression \citep{udrescu2020ai}.

Kolmogorov-Arnold Networks (KAN) \citep{liu2025kan} offer an interpretable
alternative to these neural-network architectures, since a trained KAN can be
converted into a closed-form symbolic equation that can be more easily
understood by a human. KANs have since been applied to nuclear
engineering problems, including interpretable regression for reactor
operation and safety applications \citep{panczyk2025opening,wang2025kriging}, as well as to fluid mechanics problems such as two-phase flow modeling \citep{kalesh2025physics}, turbulence anisotropy mapping \citep{kalia2025kolmogorov},
and uncertainty aware flow field reconstruction \citep{ju2026uncertainty}. These interpretable surrogates, however, have been trained almost exclusively on single-fidelity data.

KAN has recently been extended into the multifidelity setting \citep{howard2025multifidelity}. Multifidelity KAN (MFKAN) combines these two ideas: it was first introduced
as a low-fidelity KAN paired with a correction network that learns the
residual toward high-fidelity data \citep{howard2025multifidelity}. Because
a trained KAN can be converted to a symbolic equation, MFKAN also opens the
door to interpretable, closed-form multifidelity models; concurrent work has
explored applying symbolic regression directly within a KAN-based
multifidelity data-fusion framework \citep{liu2026symbolic}.

Whether KANs improve on conventional multilayer perceptrons is itself still debated: controlled comparisons on scientific problems report that some KAN variants match multilayer perceptrons in accuracy, while others are less robust to training conditions \citep{shukla2024comprehensive}. For a multifidelity KAN, the practical question is therefore whether it can match a conventional MFNN while additionally providing a symbolic model. Existing MFKAN formulations follow a residual-correction structure with two fidelity levels. To our knowledge, a single KAN conditioned directly on the fidelity level, its extension to more than two fidelity levels, and a direct comparison of MFKAN against MFNN on a large-scale engineering field have not been reported.

When generating additional high-fidelity CFD transients is computationally expensive, generative AI offers a way to augment the available data with representative synthetic samples. Generative adversarial networks have been adapted to tabular data \citep{xu2019modeling} and to time series while preserving their temporal dynamics \citep{yoon2019time}, and denoising diffusion probabilistic models \citep{ho2020denoising} underlie TabDDPM \citep{kotelnikov2023tabddpm}, a diffusion-based generative model designed for synthetic tabular data, which is used in this work. A broader overview of time-series augmentation strategies is given in \citep{wen2021time}. Similar generative data-augmentation approaches have also been explored in nuclear engineering and energy applications \citep{nabila2025data,alanazi2023multi}, supporting their potential to expand limited simulation datasets without requiring additional costly CFD calculations. Complementary strategies for limited data include uncertainty-aware variational neural networks \citep{burnett2026variational} and active learning, which reduces the number of expensive simulations by selecting only the most informative ones to run \citep{nabila2026physics}. These approaches have been developed in single-fidelity settings; their use to supply training data for multifidelity learning, when only one transient exists at each fidelity level, has not been explored.

In summary, three gaps motivate this work. First, multifidelity learning has not been applied to full-core spatiotemporal reactor fields, particularly under weak cross-fidelity correlation. Second, interpretable multifidelity models remain limited to residual-type, two-fidelity KAN formulations. Third, generative augmentation has not been combined with multifidelity learning to overcome the scarcity of CFD transients. Section~\ref{sec:methods} describes the dataset and the methods used to address these gaps.

\section{Methods}
\label{sec:methods}

\subsection{CFD Simulation and Data Generation}
\label{sec:data}

The CFD data set used in this work comes from a prior study done by our team \citep{burnett2026high}, which built a full-scale four-loop PWR model. Swirl from the centrifugal reactor coolant pumps combined with chaotic turbulent flow at the core inlet produces localized mass flow rate perturbations across the fuel assemblies that matter operationally because they can drive power tilts, giving this prediction problem safety significance. Resolving this behavior with a high-fidelity CFD transient is expensive, motivating the multifidelity approach used here.

\subsubsection{Full-Scale PWR Lower-Plenum CFD Model}
\label{sec:cfdmodel}

The model, created in STAR-CCM+, covers the four cold-leg inlets down through the
downcomer and lower plenum and ends at the core inlet, with geometry built
from the publicly available BEAVRS benchmark \citep{horelik2013benchmark}.
The 193 fuel assemblies are modeled as porous media using a Forchheimer resistance formulation, and swirl is imposed at the four cold-leg inlets through a user-defined
boundary condition to capture the non-uniform inlet conditions typical of centrifugal reactor coolant pumps.

Turbulence is modeled with the STRUCT-$\varepsilon$ closure. The complete details of the CFD model, the construction of the mesh and the formulation of turbulence are given in \citep{burnett2026high}; only the details needed to understand the machine learning tasks covered in this research are summarized here.

\subsubsection{Transient Dataset and Quantity of Interest}
\label{sec:dataset}

\begin{figure}[!ht]
\centering
\includegraphics[width=0.9\linewidth]{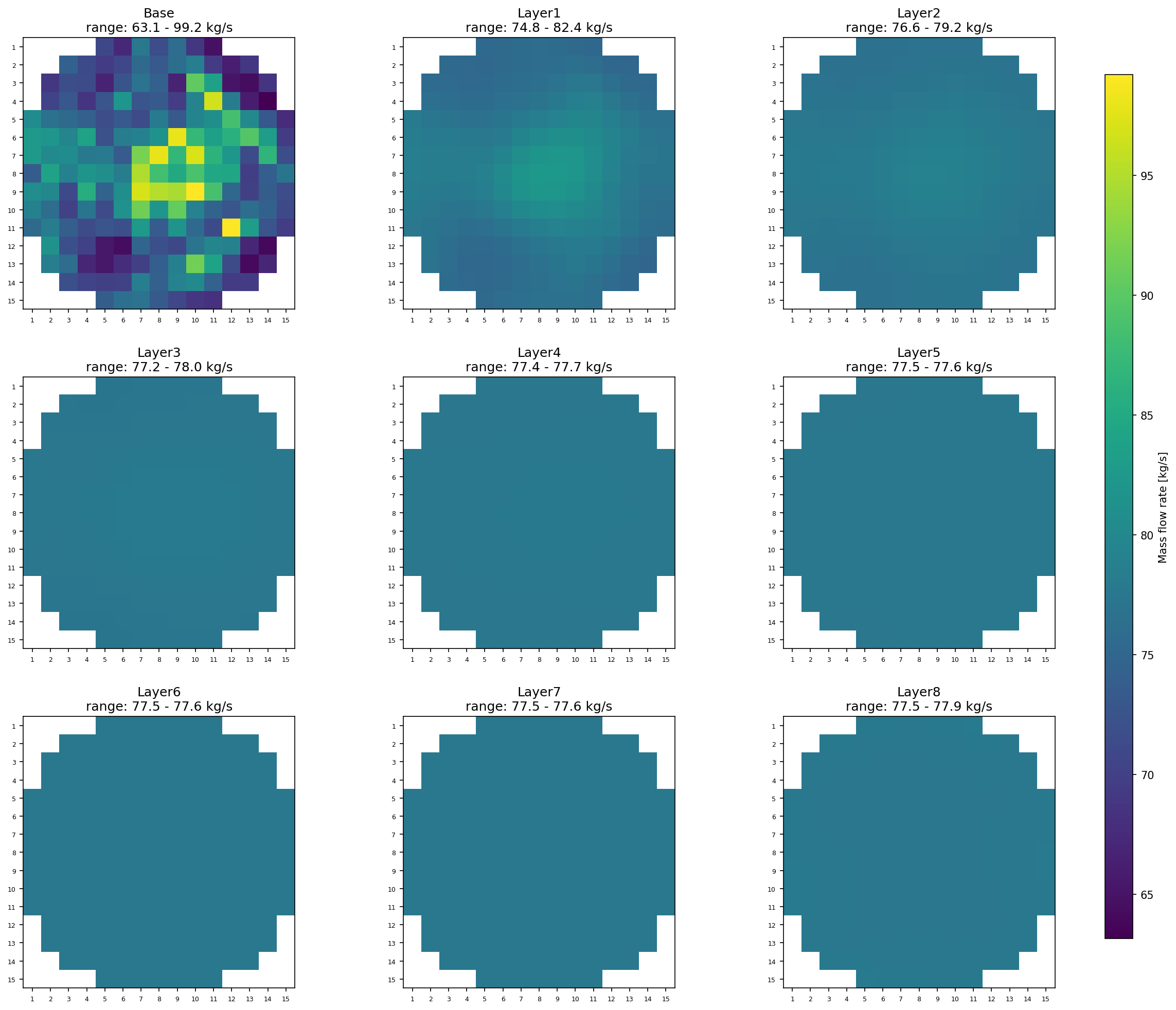}
\caption{Mass flow rate at the beginning of the usable transient across all nine axial layers, shown using a common color scale for direct comparison. \textit{Note the substantial flow heterogeneity in the base layer, which is the only layer examined further in this work}.}
\label{fig:mfr-by-layer}
\end{figure}

Each transient runs for 15 s of physical time at a fixed 0.0005 s
timestep. The first 10 s are discarded to let the flow field develop, and
data are recorded over the remaining 10 to 15 s window, giving 10,000
timesteps per transient. The recorded quantity is the area-averaged mass
flow rate at each of the 193 fuel assemblies, sampled at nine axial layers of
the CFD mesh spanning the height of the core. This paper uses only the
lowest of these nine axial layers, where the inlet flow retains the
strongest swirl-driven heterogeneity and is therefore the most chaotic and
difficult layer to predict. Figure~\ref{fig:mfr-by-layer} shows how the
mass flow rate changes axially as the flow travels upward through the
core: it varies substantially at the bottom layer, then homogenizes
quickly higher in the core. \textit{We hypothesize that flow prediction is inherently more challenging in the base layer, consistent with the findings of \cite{burnett2026high}, where the base-layer models exhibited substantially higher prediction errors than the upper layers for both flow-field reconstruction and one-step-ahead forecasting.}

\subsubsection{Fidelity Level Definition}
\label{sec:fidelity}

Three meshes at different resolutions, low, medium, and high-fidelity,
are built by scaling the fuel-assembly pitch element size to base sizes
of 2.20 cm, 1.83 cm, and 1.57 cm, respectively. Everything else about the
model, the geometry, boundary conditions, and solver settings, stays the
same across all three settings. \textit{These three meshes, denoted FApitch10, FApitch12,
and FApitch14 in the source CFD study~\cite{burnett2026high}, are treated as the LF (low-fidelity), MF (medium-fidelity), and HF (high-fidelity)
datasets used throughout this work}. Running the high-fidelity mesh on 200
cores took roughly \textbf{20 days}, and the low-fidelity mesh was done in about \textbf{6
days} on the same hardware, a gap that makes clear just how costly the
high-fidelity transient is by comparison. Only one transient is run at
each fidelity level due to substantial computational constraints.

\subsection{One-Time-Step-Forward Prediction Framing}
\label{sec:one-step}

This transient is chaotic and highly oscillatory due to turbulence, with limited temporal dependence from one timestep to the next, making long-term forecasting difficult. Similar to the temporal methods used in \citep{burnett2026high}, this work formulates prediction by pairing the state at time $t$ with the state at time $t+1$. However, static methods are used instead of sequence-based architectures, such that each input-output pair is treated as an independent training example rather than as part of a longer temporal sequence. We hypothesize that recurrent methods, such as long short-term memory (LSTM) variants, would provide little additional benefit for this problem because of the weak dependence on previous timesteps, while introducing greater network complexity and training cost compared with static methods.

Both the input and the target are 193-dimensional vectors, one value
per fuel assembly at the bottom axial layer described in
Section~\ref{sec:dataset}. At test time, the input at each step is the
true state from the CFD transient rather than a state predicted by the
model at a previous step, so error does not compound the way it would
in a genuine multi-step rollout.

Training and test sets are split chronologically rather than shuffled,
so no information from later in the transient leaks into training.
High-fidelity data, the actual prediction target, uses the first
20\% of its transient for training and the remaining 80\% for testing.
Low-fidelity data (and medium-fidelity data, for the three-fidelity
methods) is used in full for training, since it only serves as
auxiliary input rather than the quantity being evaluated. 

\textit{This one-time-step-forward framing is used unchanged for the experiments
described in Section~\ref{sec:mf-methods} and
Section~\ref{sec:multifidelity-methods}. It does not apply to the
TabDDPM-augmented forecasting direction described in
Section~\ref{sec:synthetic-augmented}, which is a genuine multi-step
forecasting task.}

\subsection{Principal Component Analysis (PCA)}
\label{sec:pca}

Principal component analysis (PCA) is a dimensionality
reduction technique. The methods in this work operate on
193-dimensional assembly-level inputs and outputs, which is too wide to
train efficiently and, for the KAN-based methods, interpretably: both KAN
parameter count and symbolic regression complexity grow quickly with
input/output size. PCA is therefore used as a preprocessing step, fit on
the high-fidelity data and applied to compress both the input and the
target before training, and as a postprocessing step, inverse transforming
predictions back into physical, per-assembly units after inference. Both
MFNN and MFKAN use PCA with the same convention. The number of components
used varies by experiment and is reported alongside the corresponding
results (Section~\ref{sec:results}). Because training is performed in PCA-compressed space, the symbolic-regression equations extracted from KAN (Sections~\ref{sec:kan} and \ref{sec:mfkan}) are expressed in terms of PCA components rather than the raw physical variables, allowing the resulting equations to remain compact. However, the analyst can still provide inputs and obtain predictions in the original physical space by applying the pre-computed PCA transformation and inverse transformation parameters.

\subsection{Cross-Fidelity Correlation Analysis}
\label{sec:correlation-analysis}

Cross-fidelity correlation refers to how strongly the mass flow rate at one mesh
resolution tracks mass flow rate at another resolution, and it matters
here because conventional multifidelity methods generally perform well
when fidelities are strongly correlated \citep{fernandez2016review}. Visual inspection of the
LF, MF, and HF transients suggested weak correlation between
fidelities, so Pearson correlation
\citep{benesty2009pearson}, was used to check that impression:
correlation is computed per fuel assembly $i$ for each fidelity pair
(LF-HF, MF-HF, LF-MF), giving per-assembly correlations $r_i$ for
$i = 1, \dots, 193$. Whole-core aggregate statistics are then reported
as the minimum, mean, and maximum of these 193 per-assembly correlation
values, i.e.\ $\min_i r_i$, $\frac{1}{193}\sum_{i=1}^{193} r_i$, and
$\max_i r_i$.

\subsubsection{Correlation-Based Assembly Subset}
\label{sec:correlated-subset-method}
The correlated-assembly subset selects the 50 assemblies with the highest
positive LF-HF Pearson correlation (Section~\ref{sec:correlation-analysis}),
and restricts both the input and output of the MFNN/MFKAN methods
(Section~\ref{sec:mf-methods}) to this subset, PCA-compressed to 10
components, rather than the full 193-assembly core. This produces a
smaller, more self-correlated prediction task that is not directly
comparable to the full-core baseline (Section~\ref{sec:results-sf-vs-mf}),
though it directly tests whether stronger cross-fidelity correlation
improves multifidelity prediction accuracy, complementing rather than
replacing the full-core baseline.
This subset also underlies the separate assembly-level analysis in
Section~\ref{sec:results-assembly-analysis}, which checks whether
TabDDPM's best-forecasting assemblies coincide with the most-correlated
ones.

\subsection{Spectral Clustering}
\label{sec:clustering}

Spectral clustering is a graph-based clustering approach that identifies groups from pairwise similarities rather than directly partitioning points in their original feature space \citep{von2007tutorial}. In the normalized spectral clustering algorithm of Ng, Jordan, and Weiss \cite{ng2002spectral}, the data are represented by a weighted similarity graph, from which the symmetric normalized graph Laplacian, $L_{\mathrm{sym}}$, is constructed. The first $k$ eigenvectors corresponding to its smallest eigenvalues form a spectral embedding, whose rows are normalized to unit length and subsequently partitioned using $k$-means clustering. This transformation provides a representation in which the underlying cluster structure can become more readily separable, including for data with complex or non-convex structure.

Clusters are formed from HF transients using the spectral clustering approach described above, producing four clusters. Each network's input still sees the full 193-assembly LF (and MF, for three-fidelity methods) core, while its output is restricted to its own cluster's assemblies, so four separate networks are trained per method. Results are reported in Section~\ref{sec:results-clustering}.

\subsection{Machine Learning Methods}
\label{sec:mf-methods}

\subsubsection{Feedforward Neural Networks (FNN)}
\label{sec:fnn}

A feedforward neural network (FNN) \citep{goodfellow2016deep}
approximates a function by composing layers of nodes, where each node
applies a fixed nonlinear activation function to a weighted sum of its
inputs from the previous layer. The weights and biases are learned via
backpropagation to minimize a loss function on the training data. FNNs
serve as the standard general-purpose function approximator used here
and form the building block for the multifidelity FNN baseline described
in Section~\ref{sec:mfnn}.

\subsubsection{Kolmogorov-Arnold Networks (KAN)}
\label{sec:kan}

A Kolmogorov-Arnold Network (KAN) \citep{liu2025kan} differs from a
standard feedforward neural network (FNN) in how nonlinear
transformations are represented and learned. In an FNN, connections are
parameterized by trainable scalar weights, while fixed nonlinear
activation functions are applied at the nodes. In contrast, KAN replaces
these scalar weights with trainable univariate functions, typically
parameterized using splines, while each node sums the outputs of its
incoming functions. Thus, nonlinear transformations are learned directly
on the network connections rather than imposed through fixed node
activation functions. Because these learned functions are explicit and
univariate, they can be inspected, simplified, and approximated by
analytical functions after training, allowing the trained network to be
expressed as a closed-form symbolic model through symbolic regression.
This enhanced interpretability motivates the use of KAN throughout this
work and forms the basis for the multifidelity KAN formulations
introduced in Section~\ref{sec:mfkan}.

After training, the learned KAN spline functions can be approximated by
closed-form symbolic expressions using pykan's
\texttt{auto\_symbolic()} procedure. This procedure fits a library of
candidate elementary functions to each trained spline and replaces the
spline with the best-matching symbolic form. Symbolic
regression is performed on the same PCA-compressed inputs and outputs
used during training (Section~\ref{sec:pca}), rather than on the raw
physical variables. Throughout this work, the resulting symbolic model is evaluated on the same test set as the original spline-based model, allowing their predictive accuracies to be
compared directly.

\subsection{Multifidelity Methods}
\label{sec:multifidelity-methods}

\subsubsection{Multifidelity FNN (MFNN)}
\label{sec:mfnn}

Three multifidelity formulations were investigated, following the same Delta and Flag implementations and structures used in our prior work \citep{eaheart2026multifidelity}, adapted here as neural-network baselines for the MFKAN comparison in Section~\ref{sec:mfkan}. These baselines establish whether conventional multifidelity learning can exploit whatever cross-fidelity structure is present in this dataset (Section~\ref{sec:correlation-analysis}).

\paragraph{MFNN: Delta Method}
The Delta method~\cite{meng2020composite}  models the high-fidelity response as a correction to a learned low-fidelity prediction. First, an FNN $f_L(x)$ is trained on the low-fidelity dataset. The trained model is evaluated at the high-fidelity input locations, and the discrepancy between the high-fidelity observations and these predictions is computed as
\begin{equation}
    \label{eq:delta-residual}
    r(x) = y_H(x) - f_L(x)
\end{equation}
where $x$ denotes the high-fidelity input locations. A second FNN $f_{\Delta}(x, f_L(x))$ is then trained to model this residual at those same high-fidelity locations, and the final prediction is
\begin{equation}
    \label{eq:delta-combine}
    \hat{y}_H(x) = f_L(x) + f_{\Delta}(x, f_L(x))
\end{equation}
The two networks are trained sequentially and do not require the low- and high-fidelity datasets to share identical sampling locations.

\paragraph{MFNN: Flag Method}
In the fidelity-flag approach~\cite{guo2022multi}, a single FNN is trained to learn multiple fidelity mappings simultaneously. The fidelity level is incorporated as an additional input feature by appending a fidelity indicator $f$ to the input vector, with $f=0$ denoting low-fidelity and $f=1$ denoting high-fidelity data. High-fidelity predictions are obtained by evaluating the learned mapping $F(x,f)$ at $f=1$,
\begin{equation}
    \hat{y}_H(x) = F(x,1)
\end{equation}
Correlations between fidelity levels are learned implicitly through the network's shared parameters.

\paragraph{MFNN: Three-Fidelity Flag Method}
The Flag method extends naturally to three fidelity levels by generalizing the fidelity indicator to $f \in \{0,1,2\}$ for low-, medium-, and high-fidelity data, respectively. High-fidelity predictions are obtained by evaluating $F(x,f)$ at $f=2$,
\begin{equation}
    \hat{y}_H(x) = F(x,2)
\end{equation}

\subsubsection{Multifidelity KAN (MFKAN)}
\label{sec:mfkan}

Three multifidelity KAN formulations were investigated, following the same Delta and Flag structures described for MFNN above (Section~\ref{sec:mfnn}), with each FNN replaced by a KAN. After an initial fit, each KAN is pruned, then retrained, following pykan's standard training convention.

\paragraph{MFKAN: Delta Method}
The Delta method follows the same construction as MFNN-Delta above (Section~\ref{sec:mfnn}, Eqs.~\eqref{eq:delta-residual}--\eqref{eq:delta-combine}), with $f_L$ and $f_{\Delta}$ implemented as KANs instead of FNNs.

Symbolic regression (Section~\ref{sec:kan}) is applied independently to the trained $f_L$ and $f_{\Delta}$ KANs, converting each network's learned splines into its own closed-form symbolic expression, $f_L^{\text{sym}}$ and $f_{\Delta}^{\text{sym}}$ respectively. These two symbolic models are then chained the same way as their numeric forward pass to produce the final symbolic prediction:
\begin{equation}
    \hat{y}_H^{\text{sym}}(x) = f_L^{\text{sym}}(x) + f_{\Delta}^{\text{sym}}\big(x, f_L^{\text{sym}}(x)\big).
\end{equation}
That is, $f_L$'s symbolic expression is evaluated first, and its output is fed into $f_{\Delta}$'s symbolic expression as part of its input.

\paragraph{MFKAN: Flag Method (Proposed)}
The Flag method follows the same construction as MFNN-Flag above (Section~\ref{sec:mfnn}), with $F$ implemented as a single KAN instead of an FNN, which we propose here as a new MFKAN variant. Symbolic regression (Section~\ref{sec:kan}) is applied once to the single trained KAN, producing one symbolic model with the fidelity flag as an input variable alongside the PCA-compressed inputs (Section~\ref{sec:pca}). Only one lossy spline-to-symbolic conversion is needed here, unlike MFKAN-Delta's two symbolic conversions.

\paragraph{MFKAN: Three-Fidelity Flag Method (Proposed)}
The three-fidelity Flag method, which we call \textit{MFKAN-Flag3f}, extends the two-fidelity MFKAN-Flag method proposed above to three fidelity levels, following the same construction as MFNN's three-fidelity Flag method above (Section~\ref{sec:mfnn}), with $F$ implemented as a single KAN instead of an FNN. As with MFKAN-Flag, symbolic regression is applied once to the single trained KAN, with the fidelity flag as an input variable alongside the PCA-compressed inputs.

\subsubsection{Implementation Validation on Analytical Benchmarks}
\label{sec:mfkan-benchmarks}

Before being applied to the reactor CFD dataset, the MFKAN-Delta and MFKAN-Flag
implementations were validated on the same two-fidelity analytical
benchmark functions used in \citep{eaheart2026multifidelity}: the
bi-fidelity Forrester function \citep{forrester2007multi} and the
bi-fidelity Branin function, both drawn from the MF2 package
\citep{van2020mf2}. MFKAN-Flag3f was similarly validated on
three-fidelity Forrester (1D), Rastrigin (2D), and Rosenbrock (2D)
functions \citep{mainini2022analytical}. These benchmarks have known analytical
solutions, so they confirm each implementation reproduces sensible
multifidelity behavior before being applied to the real-world engineering dataset, which
has no such ground truth to check against directly.

\paragraph{Bi-fidelity Forrester Function (1D)}

The bi-fidelity Forrester function is derived from the classical one-dimensional Forrester test function~\cite{forrester2007multi} as shown below:
\begin{align}
f_h(x) &= (6x - 2)^2 \sin(12x - 4) \\
f_l(x) &= A f_h(x) + B(x - 0.5) + C
\end{align}
where $x \in [0,1]$. The recommended parameters are:
\[
A = 0.5, \quad B = 10, \quad C = -5
\]

\paragraph{Bi-fidelity Branin Function (2D)}
The bi-fidelity Branin function is defined by~\cite{van2020mf2} as
\begin{align}
    f_b(x_1, x_2) &= \left(x_2 - 5.1\frac{x_1^2}{4\pi^2} + \frac{5x_1}{\pi} - 6\right)^2 + 10\cos(x_1)\left(1 - \frac{1}{8\pi}\right) + 10, \\
    f_h(x_1, x_2) &= f_b(x_1, x_2) - 22.5x_2, \\
    f_l(x_1, x_2) &= f_b(0.7x_1, 0.7x_2) - 15.75x_2 + 20(0.9+x_1)^2 - 50,
\end{align}
where $x_1 \in [-5,10]$ and $x_2 \in [0,15]$.

\paragraph{Multi-fidelity Forrester (Adapted to Tri-Fidelity)}
The multifidelity Forrester function is defined as shown below~\cite{mainini2022analytical,forrester2007multi}:
 \begin{align}
f_1(x) &= (6x - 2)^2 \sin(12x - 4), \\[6pt]
f_2(x) &= (5.5x - 2.5)^2 \sin(12x - 4), \\[6pt]
f_3(x) &= 0.75(6x - 2)^2 \sin(12x - 4) + 5(x - 0.5) - 2, \\[6pt]
f_4(x) &= 0.5(6x - 2)^2 \sin(12x - 4) + 10(x - 0.5) - 5,
\end{align}
where $x \in [0,1]$. This formulation defines four fidelity levels, ordered from highest ($f_1$) to lowest ($f_4$), but only three are needed here. $f_1$ is therefore omitted, and the remaining three are reassigned as high-fidelity ($f_2$), medium-fidelity ($f_3$), and low-fidelity ($f_4$).

\paragraph{Tri-fidelity Rosenbrock} 
The tri-fidelity Rosenbrock function is defined as shown below~\cite{mainini2022analytical,bryson2016variable,ficini2021assessing}:
\begin{equation}
f_1(\mathbf{x}) =
\sum_{i=1}^{D-1} \left[ 100\left(x_{i+1} - x_i^2\right)^2 + (1 - x_i)^2 \right]
\end{equation}

\begin{equation}
f_2(\mathbf{x}) = \sum_{i=1}^{D-1}\left[50\left(x_{i+1} - x_i^2\right)^2 + (-2 - x_i)^2\right] - \sum_{i=1}^{D} 0.5\,x_i
\end{equation}

\begin{equation}
f_3(\mathbf{x}) =
\frac{f_1(\mathbf{x}) - 4 - \sum_{i=1}^{D} 0.5\,x_i}
     {10 + \sum_{i=1}^{D} 0.25\,x_i}
\end{equation}

\[
\mathbf{x} \in [-2,2]^D
\]

Similarly to the representation in the Forrester function, $f_3$ represents the lowest fidelity, $f_2$ is the medium-fidelity, and $f_1$ is the high-fidelity solution. One interesting feature of these equations is that the number of input dimensions can be changed by refining the $D$ parameter. In this work, $D=2$ was used to validate the performance of the three-fidelity methods.

\paragraph{Tri-fidelity Rastrigin}
The tri-fidelity shifted-rotated Rastrigin function~\cite{ficini2021assessing,wang2017generic} is defined as

\begin{equation}
f_i(\mathbf{z};\phi_i) = f_1(\mathbf{z}) + e_r(\mathbf{z}, \phi_i), \quad i=1,2,3,
\end{equation}
where the base Rastrigin function is
\begin{equation}
f_1(\mathbf{z}) = \sum_{j=1}^{D} \left( z_j^2 + 1 - \cos(10\pi z_j) \right),
\end{equation}
and
\begin{equation}
\mathbf{z} = R_D(\theta)(\mathbf{x} - \mathbf{x}^*).
\end{equation}

For the two-dimensional case used in this work,
\begin{equation}
R_2(\theta)=
\begin{pmatrix}
\cos\theta & -\sin\theta\\
\sin\theta & \cos\theta
\end{pmatrix}.
\end{equation}
The fidelity-dependent error term is
\begin{equation}
e_r(\mathbf{z}, \phi) =
\sum_{j=1}^{D} a(\phi)\cos^2\!\left(w(\phi) z_j + b(\phi) + \pi \right),
\end{equation}
with
\begin{align}
a(\phi) &= \Theta(\phi),\\
w(\phi) &= 10\pi\Theta(\phi),\\
b(\phi) &= 0.5\pi\Theta(\phi),\\
\Theta(\phi) &= 1 - 0.0001\phi.
\end{align}

The input domain is
\[
-0.1 \le x_i \le 0.2, \qquad i=1,\ldots,D,
\]
the rotation angle is $\theta=0.2$, and the optimum occurs at
\[
\mathbf{x}^* = (0.1,\ldots,0.1)^\top,
\]
for which the highest-fidelity objective value is zero. The fidelity parameters are
\[
\phi_1 = 10000 \quad \text{(high-fidelity)}, \qquad
\phi_2 = 5000 \quad \text{(medium-fidelity)}, \qquad
\phi_3 = 2500 \quad \text{(low-fidelity)}.
\]

\subsection{Generative Models for Data Augmentation}
\label{sec:synthetic}

Because only one CFD transient is currently available at each fidelity level, the limited diversity of the training data presents a major challenge for transient prediction. Generating hundreds or thousands of additional CFD transients to expand the training space is computationally prohibitive, motivating synthetic-data augmentation as a practical alternative. \textit{The last leg of this work therefore investigates whether generative models can extract the statistical structure of the available CFD data and generate additional representative samples that improve transient prediction when real training data are severely limited.}

\subsubsection{TabDDPM Method}
\label{sec:tabddpm}

TabDDPM \citep{kotelnikov2023tabddpm}, a diffusion-based generative model for tabular data, is used to generate additional synthetic transients at each fidelity level, addressing the scarcity of real CFD data (only one transient available per fidelity, Section~\ref{sec:dataset}). TabDDPM is trained once per fidelity level, then sampled repeatedly without retraining. Each newly generated sample is checked against every assembly's real trajectory and added to whichever assembly's bucket it correlated most strongly with, continuing until every assembly has accumulated 1{,}000 synthetic realizations. The computational cost of this matching process varied substantially by assembly; see Section~\ref{sec:results-tabddpm} for details.

\subsubsection{Synthetic Data Validation}
\label{sec:synthetic-validation}

Synthetic data quality is validated using the same Pearson correlation metric introduced in Section~\ref{sec:correlation-analysis}, computed here between each assembly's synthetic realizations and that assembly's own real trajectory at the same fidelity, rather than across fidelities.

\subsection{TabDDPM-Augmented Forecasting Experimental Design}
\label{sec:synthetic-augmented}

Synthetic realizations from Section~\ref{sec:tabddpm} are used to train single-fidelity KAN and FNN forecasting models (Section~\ref{sec:mf-methods}). Two types of experiments are run, both tested with $N=1$ and $N=10$: the first $N$ timesteps of the real transient being predicted are given to the model as known values, and the model must forecast the remaining timesteps on its own. In self-evaluation, a model is trained on one fidelity's synthetic data and evaluated on that same fidelity's real transient. In cross-fidelity forecasting, a model is trained on one or more fidelities' synthetic data and evaluated on a different fidelity's real transient, across three directions: LF$\to$HF, MF$\to$HF, and LF+MF$\to$HF. Symbolic regression (Section~\ref{sec:kan}) is applied separately to these forecasting-specific KAN models, which are newly trained for this cross-fidelity setting rather than reused from the one-time-step-forward models in Section~\ref{sec:mf-methods}. The resulting symbolic equations are compared against their own spline representations within this same forecasting setting, extending the spline-vs-symbolic comparison from Section~\ref{sec:results-symbolic-vs-spline} to this task.

\subsection{Hyperparameter Tuning}
\label{sec:hyperparameter-tuning}

Hyperparameters were tuned separately for MFKAN, MFNN (Sections~\ref{sec:mfnn}--\ref{sec:mfkan}), and the dedicated KAN and FNN forecasting models (Section~\ref{sec:synthetic-augmented}), using Bayesian optimization in Optuna \citep{akiba2019optuna}. MFKAN and MFNN were tuned on the one-time-step-forward baseline case (Section~\ref{sec:results-sf-vs-mf}): the 20\% high-fidelity training / 80\% testing split used throughout Section~\ref{sec:results-one-step-forward}. The KAN and FNN forecasting models were tuned on the self-evaluation setting (e.g.\ LF$\to$LF and MF$\to$MF, $N=1$; Section~\ref{sec:results-forecasting}), then applied across the full set of forecasting directions. Table~\ref{tab:hyperparameters-search} summarizes the full search space and which values were held fixed.

All four searches ran for up to 100 trials, with progress checkpointed so a search could be resumed without restarting. The two forecasting searches train on the much larger synthetic TabDDPM dataset, making each trial more expensive, so they were stopped early once improvement became marginal, and the FNN forecasting search also used a lower range of training epochs than MFNN's (Table~\ref{tab:hyperparameters-search}). The combined-fidelity forecasting models (LF+MF$\to$HF) were not tuned separately; they reuse the dedicated low-fidelity model's winning hyperparameters. Final tuned values are reported alongside their respective results, in Sections~\ref{sec:results-one-step-forward-hyperparameters} and \ref{sec:results-forecasting-hyperparameters}.

\input{tables/tab_hyperparameters_search.tex}

\section{Results}
\label{sec:results}

Results are organized into four subsections. Section~\ref{sec:results-benchmarks} covers the analytical-benchmark validation of the MFKAN implementations. Section~\ref{sec:results-correlation} covers the cross-fidelity correlation analysis, which is one of the primary drivers of multifidelity method performance and motivates the two experimental designs that follow. Section~\ref{sec:results-one-step-forward} covers the one-time-step-forward prediction method, including the tuned hyperparameters used (Section~\ref{sec:results-one-step-forward-hyperparameters}) and results on the baseline (Section~\ref{sec:results-sf-vs-mf}), clustered (Section~\ref{sec:results-clustering}), and most-correlated-subset (Section~\ref{sec:results-correlated-subset}) assembly groupings, with the symbolic regression results covered in Section \ref{sec:results-symbolic-vs-spline}. Section~\ref{sec:results-tabddpm-section} covers the TabDDPM data augmentation study, including synthetic data generation and validation (Section~\ref{sec:results-tabddpm}), the tuned hyperparameters used (Section~\ref{sec:results-forecasting-hyperparameters}), augmented forecasting results (Section~\ref{sec:results-forecasting}), symbolic regression (Section~\ref{sec:results-tabddpm-symbolic-vs-spline}), and assembly-level analysis (Section~\ref{sec:results-assembly-analysis}).

\begin{figure}[!ht]
\centering
\includegraphics[width=0.6\linewidth]{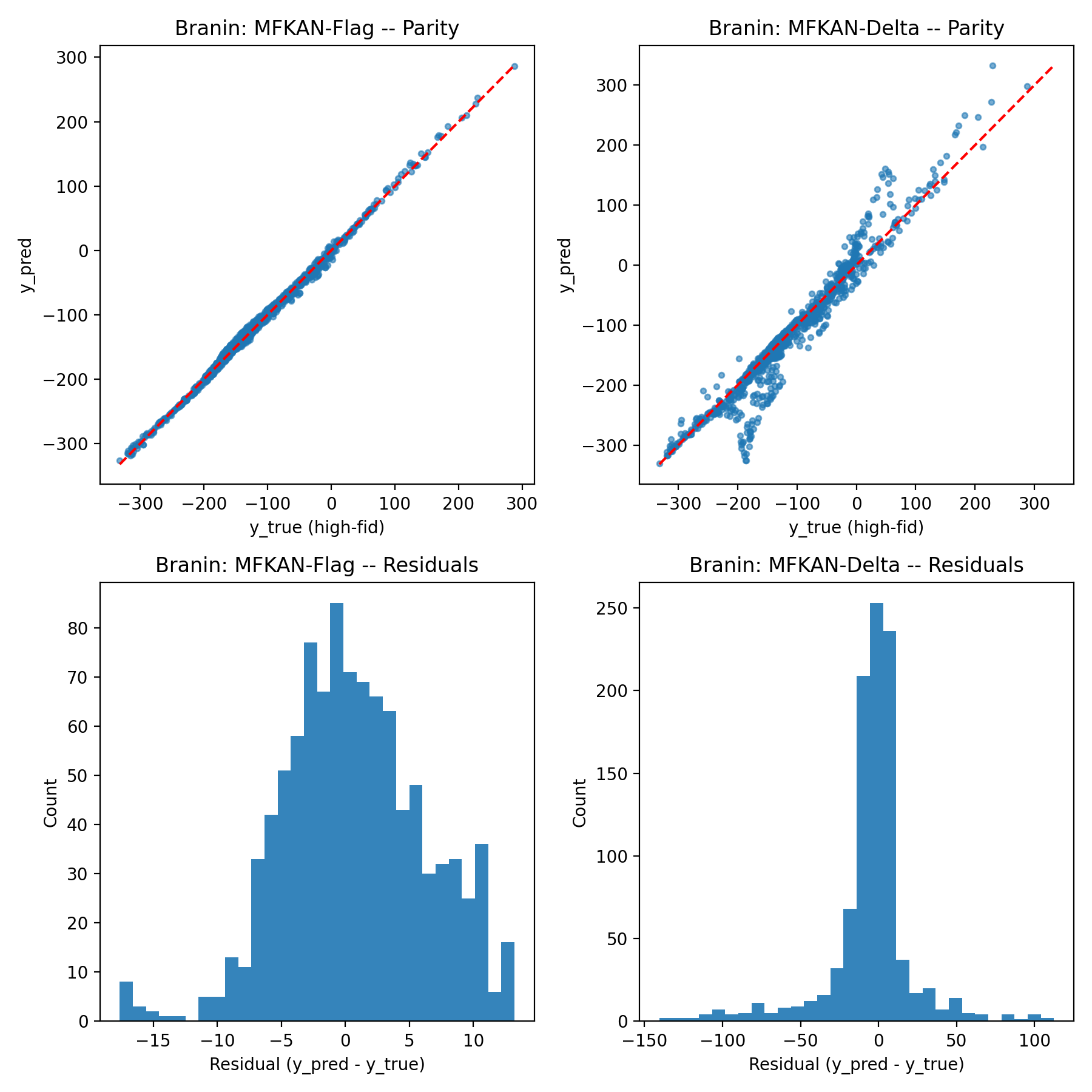}
\caption{MFKAN-Flag and MFKAN-Delta parity plots (top) and residual histograms (bottom) on the bi-fidelity Branin function.}
\label{fig:mfkan-benchmarks-2f}
\end{figure}

\subsection{Benchmark Results}
\label{sec:results-benchmarks}

\begin{figure}[!ht]
\centering
\includegraphics[width=0.8\linewidth]{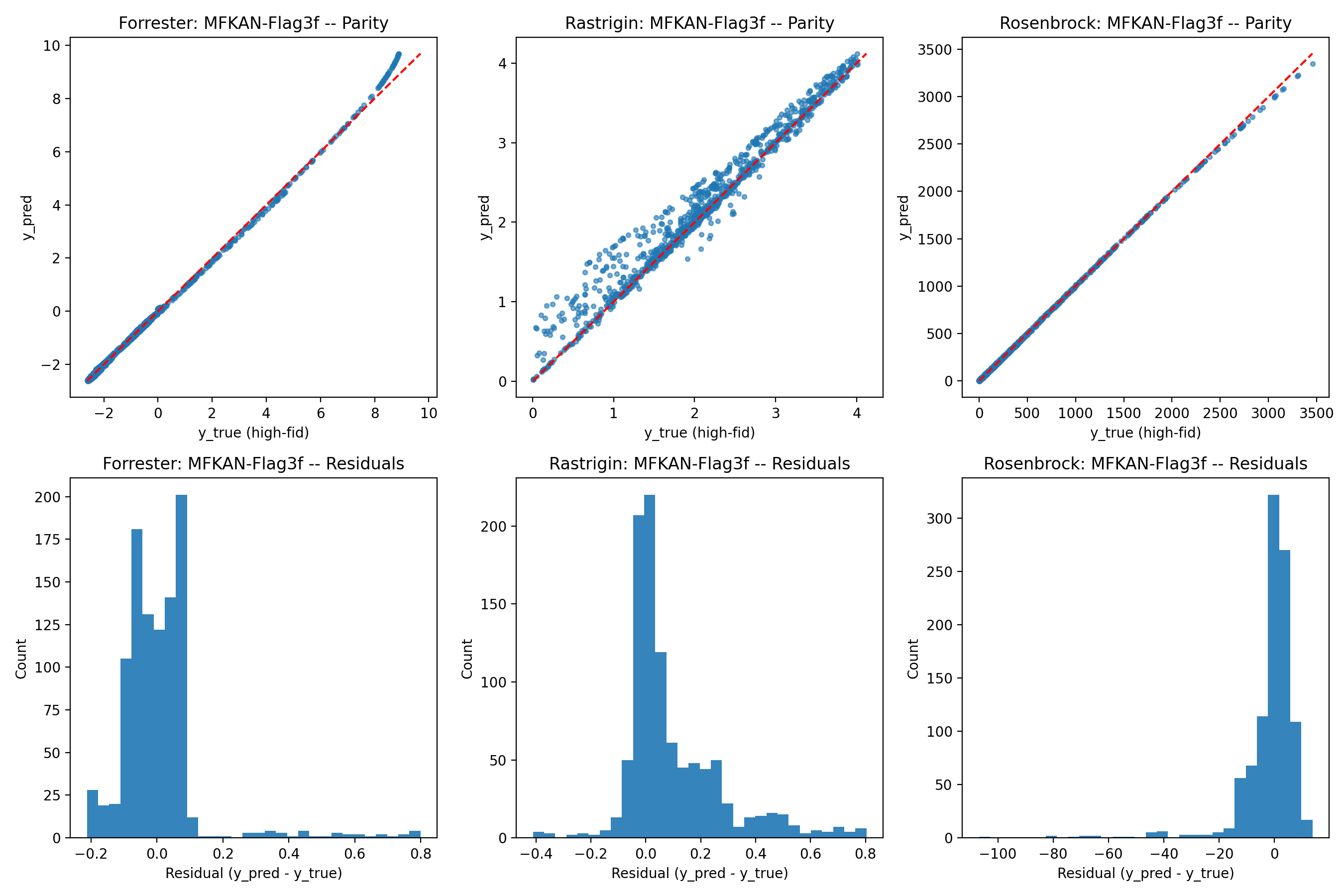}
\caption{MFKAN-Flag3f parity plots (top) and residual histograms (bottom) on the three-fidelity Forrester (1D), Rastrigin (2D), and Rosenbrock (2D) functions.}
\label{fig:mfkan-benchmarks-3f}
\end{figure}

Section~\ref{sec:mfkan-benchmarks} describes the analytical-benchmark validation performed on each MFKAN implementation before it was applied to the reactor CFD dataset. Figure~\ref{fig:mfkan-benchmarks-2f} shows parity plots and residual histograms for MFKAN-Flag and MFKAN-Delta on the bi-fidelity Branin function, and Figure~\ref{fig:mfkan-benchmarks-3f} shows the same for MFKAN-Flag3f on the three-fidelity Forrester (1D), Rastrigin (2D), and Rosenbrock (2D) functions. The five method/benchmark combinations shown span $R^2 = 0.908$ to $0.9997$, and the bi-fidelity Forrester case (not plotted for brevity, since both methods fit it near-perfectly) reaches $R^2 = 0.9997$ for MFKAN-Flag and $R^2 = 0.9999$ for MFKAN-Delta. MFKAN-Delta on Branin is the weakest of the five ($R^2 = 0.908$), visibly noisier than the other combinations, possibly reflecting error accumulation across its two sequentially trained networks, whereas MFKAN-Flag learns both fidelities within a single network. No extensive hyperparameter search was performed in this section, as the objective was to demonstrate proof of concept and verify the implementation rather than optimize predictive performance. Together, these results confirm that each implementation reproduces expected multifidelity behavior on problems with a known analytical solution, before being applied to the reactor CFD dataset, which has no such ground truth to check against directly.

\subsection{Cross-Fidelity Correlation Analysis Results}
\label{sec:results-correlation}

This section presents the cross-fidelity correlation analysis for the LF-HF and MF-HF assembly pairs across the full 193-assembly core. The data varies substantially between fidelities, which makes this a difficult problem for multifidelity machine learning methods. Pearson correlation is used as a metric throughout the paper, alongside the more standard mean absolute error (MAE) and root mean squared error (RMSE), because it captures how well a method tracks the overall shape of the transient in a way those other metrics do not. A method can score a high RMSE while still having a strong correlation, since the sharp flow rate peaks in this dataset are hard to capture exactly, so the predicted transient can track the shape correctly without ever reaching the full height of the peak.

\begin{figure}[!ht]
\centering
\includegraphics[width=0.75\linewidth]{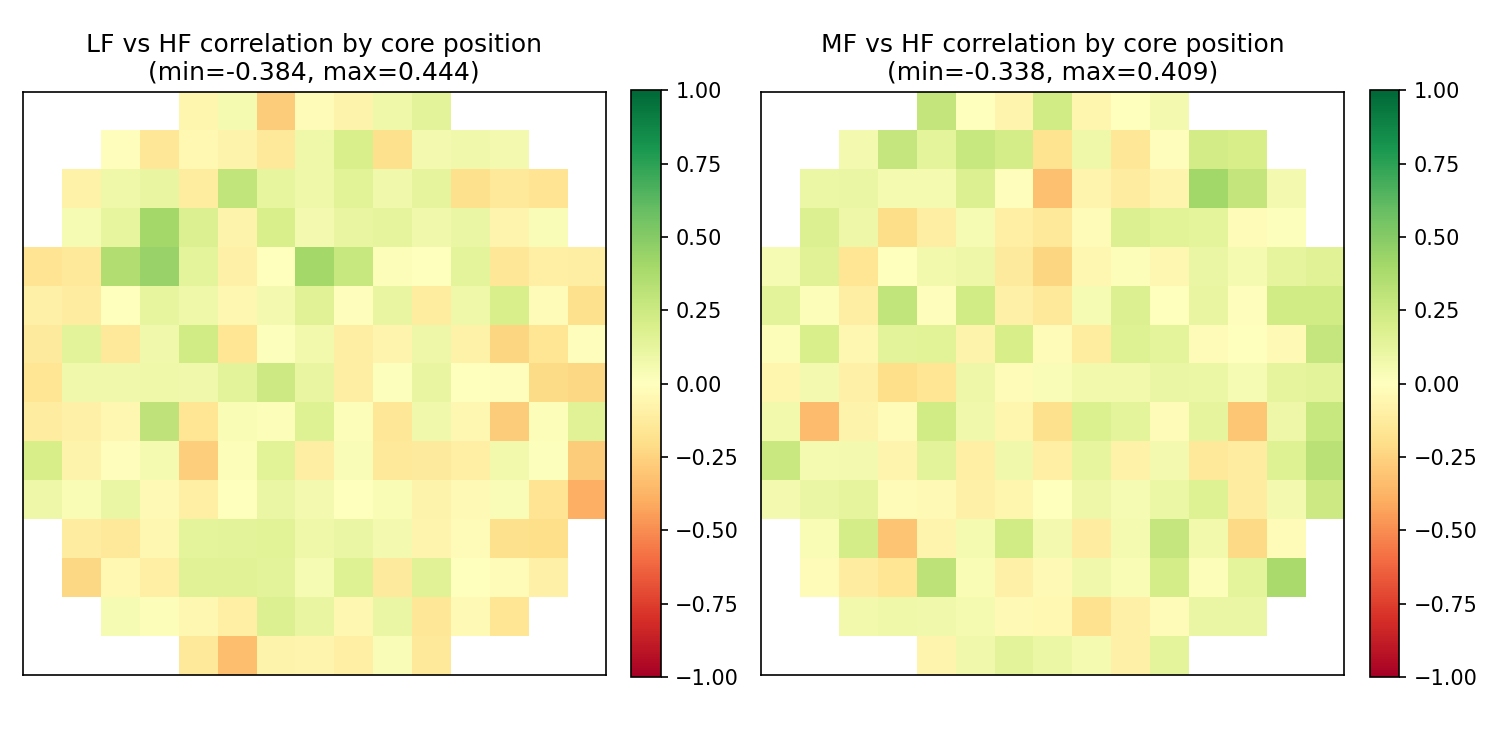}
\caption{LF-HF and MF-HF correlation, computed between the real transients at each fidelity, mapped onto the core layout.}
\label{fig:corr-core-heatmap}
\end{figure}

Correlation across the core can be seen in Figure~\ref{fig:corr-core-heatmap}, which maps LF-HF and MF-HF correlation onto the core layout. This correlation is computed directly between the real transients recorded at each fidelity level during the original CFD simulations (Section~\ref{sec:correlation-analysis}), not between any trained model's predictions and ground truth. No MFNN or MFKAN model is involved in this analysis. Across all 193 assemblies, the mean LF-HF correlation is only 0.003 (median 0.008) and the mean MF-HF correlation is 0.041 (median 0.053), both essentially zero. Even the single best assembly only reaches 0.444 for LF-HF and 0.409 for MF-HF. The two heatmaps show different spatial patterns, indicating that the cross-fidelity correlation is not uniformly weak everywhere: it varies by assembly and by which fidelity pair is being compared, so a location that is highly correlated LF-to-HF is not necessarily the same location that is highly correlated MF-to-HF.

\begin{figure}[!ht]
\centering
\includegraphics[width=0.75\linewidth]{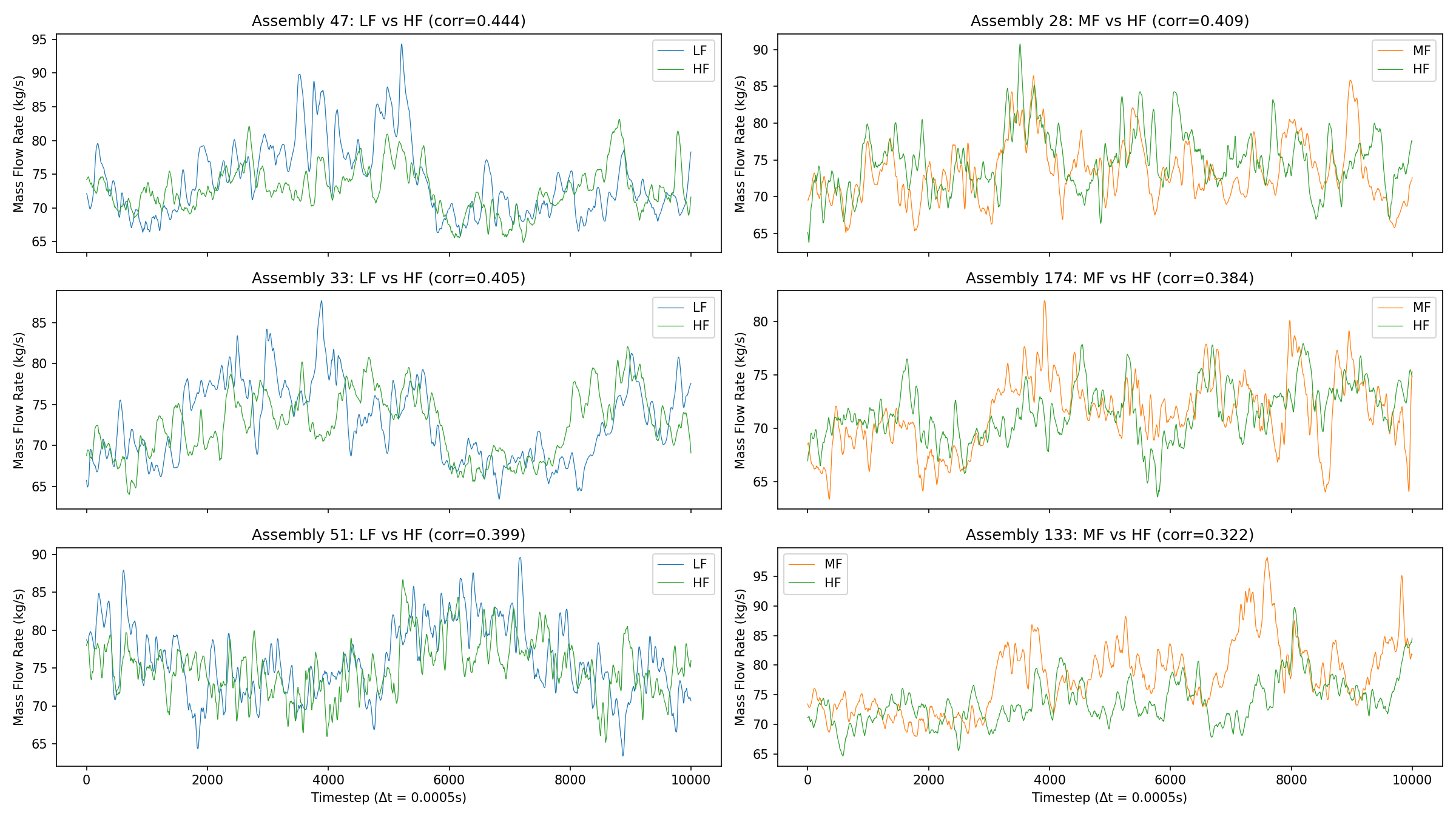}
\caption{Mass flow rate trajectories for the \textbf{highest-correlation assemblies}, with LF--HF and MF--HF pairs ranked independently. Even the highest-correlation cases exhibit only moderate agreement.}
\label{fig:corr-timeseries-most}
\end{figure}

\begin{figure}[!ht]
\centering
\includegraphics[width=0.75\linewidth]{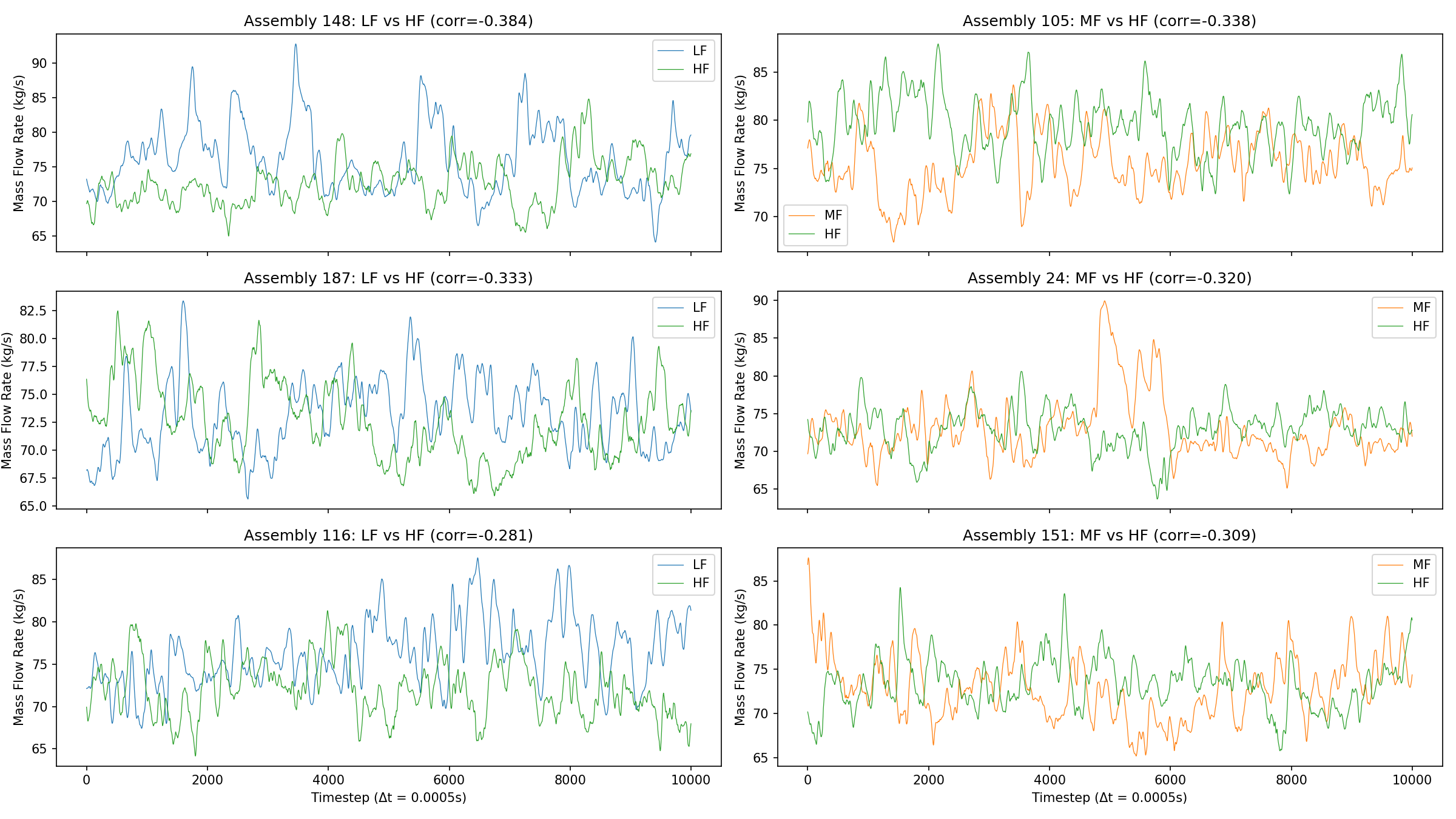}
\caption{Mass flow rate trajectories for the \textbf{lowest-correlation assemblies}, with LF--HF and MF--HF pairs ranked independently. These cases exhibit weak cross-fidelity agreement.}
\label{fig:corr-timeseries-least}
\end{figure}

To illustrate the behavior of the most- and least-correlated assemblies, Figures~\ref{fig:corr-timeseries-most} and~\ref{fig:corr-timeseries-least} compare their mass flow rate trajectories across fidelity levels. Even the most-correlated assemblies exhibit only moderate agreement, highlighting the difficulty of this multifidelity problem. For the least-correlated assemblies, the trajectories differ substantially, providing limited cross-fidelity information for predicting the high-fidelity transient from low- or medium-fidelity data.

Overall, cross-fidelity correlation remains weak across the core, motivating the central hypothesis examined in Section~\ref{sec:results-one-step-forward}: that the predictive performance of MFNN and MFKAN is fundamentally constrained by the limited correlation between the low- and medium-fidelity data and the high-fidelity ground truth.

\subsection{One-Time-Step-Forward Prediction Results}
\label{sec:results-one-step-forward}
These results correspond to the one-time-step-ahead methodology, in which all low-fidelity data (and medium-fidelity data for the three-fidelity methods) are used for training together with 20\% of the high-fidelity data. The remaining 80\% of the high-fidelity transient is then predicted one time step ahead and compared with the ground truth. This approach can be considered one-step-ahead forecasting because each prediction advances beyond the state provided to the model; however, the errors do not compound over time because each prediction uses the true input state rather than feeding the previous prediction back as the next input. \textit{This differs from the TabDDPM-based forecasting approach presented in Section~\ref{sec:results-tabddpm-section}, where predictions are propagated forward recursively}.

Three experiments are considered in this section: a baseline using all assemblies without grouping or subsetting, a spectral-clustering approach based on Section \ref{sec:clustering}, and a subset containing the most highly correlated assemblies.

\subsubsection{Hyperparameter Tuning Results}
\label{sec:results-one-step-forward-hyperparameters}

Final tuned hyperparameters for the MFKAN and MFNN baseline methods (search space and tuning methodology in Section~\ref{sec:hyperparameter-tuning}) are reported in Tables~\ref{tab:hyperparameters-mfkan} and \ref{tab:hyperparameters-mfnn}.

\input{tables/tab_hyperparameters_mfkan.tex}
\input{tables/tab_hyperparameters_mfnn.tex}

\subsubsection{Baseline}
\label{sec:results-sf-vs-mf}
This is the baseline case: all 193 assemblies are used together as a single group, with no clustering and no correlated-assembly subsetting. Inputs and outputs are PCA-compressed to 10 components (Section~\ref{sec:pca}) before training the MFNNs and MFKANs.

\begin{figure}[!ht]
\centering
\includegraphics[width=0.9\linewidth]{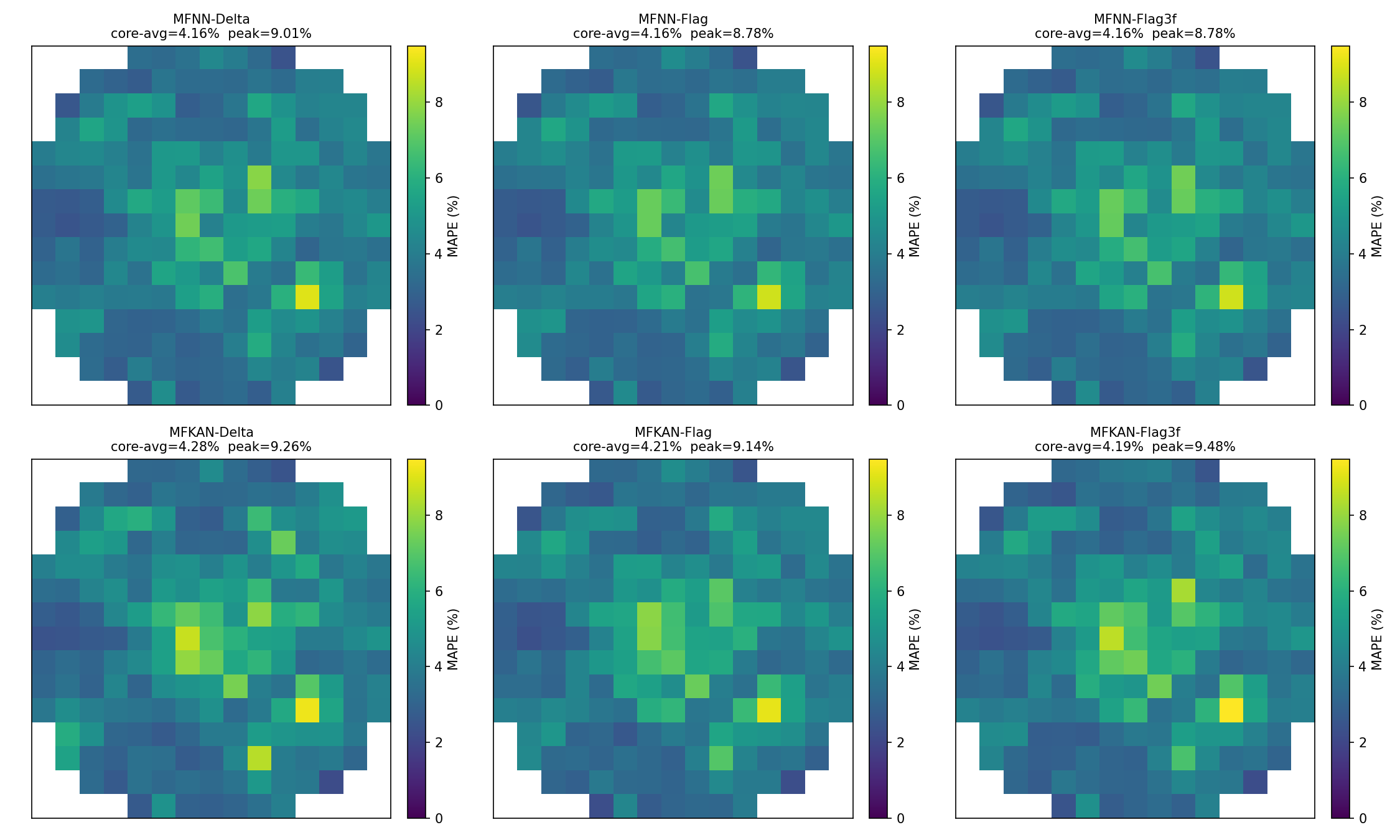}
\caption{Assembly-wise MAPE distributions for all six multifidelity methods using 20\% of the high-fidelity data for training in the \textbf{baseline} case. Core-average MAPE and peak assembly MAPE are reported for each method.}
\label{fig:real-data-mape-comparison}
\end{figure}

As shown in Figure~\ref{fig:real-data-mape-comparison}, all six methods performed similarly, with the MFNN methods slightly outperforming the MFKAN methods. The lowest core-average MAPE is tied between MFNN-Delta, MFNN-Flag, and MFNN-Flag3f at 4.16\%, and the lowest peak MAPE is tied between MFNN-Flag and MFNN-Flag3f at 8.78\%. The MFKAN methods performed similarly to the MFNN methods overall, with the added benefit of interpretability.

A constant random seed was used, but repeated runs still showed some variability, prompting the reproducibility study in Table~\ref{tab:reproducibility}. MFKAN-Delta showed the most run-to-run variability of the six methods, consistent with the random-seed sensitivity reported for KAN generally \cite{liu2025kan}; MFNN was highly stable, and MFKAN-Flag/Flag3f varied only slightly. Aside from this reproducibility study itself, a single MFKAN-Delta run is used throughout the rest of the paper, so its results may vary slightly on rerun.

\input{tables/tab_reproducibility.tex}

\subsubsection{Clustered}
\label{sec:results-clustering}

\begin{figure}[!h]
\centering
\includegraphics[width=0.9\linewidth]{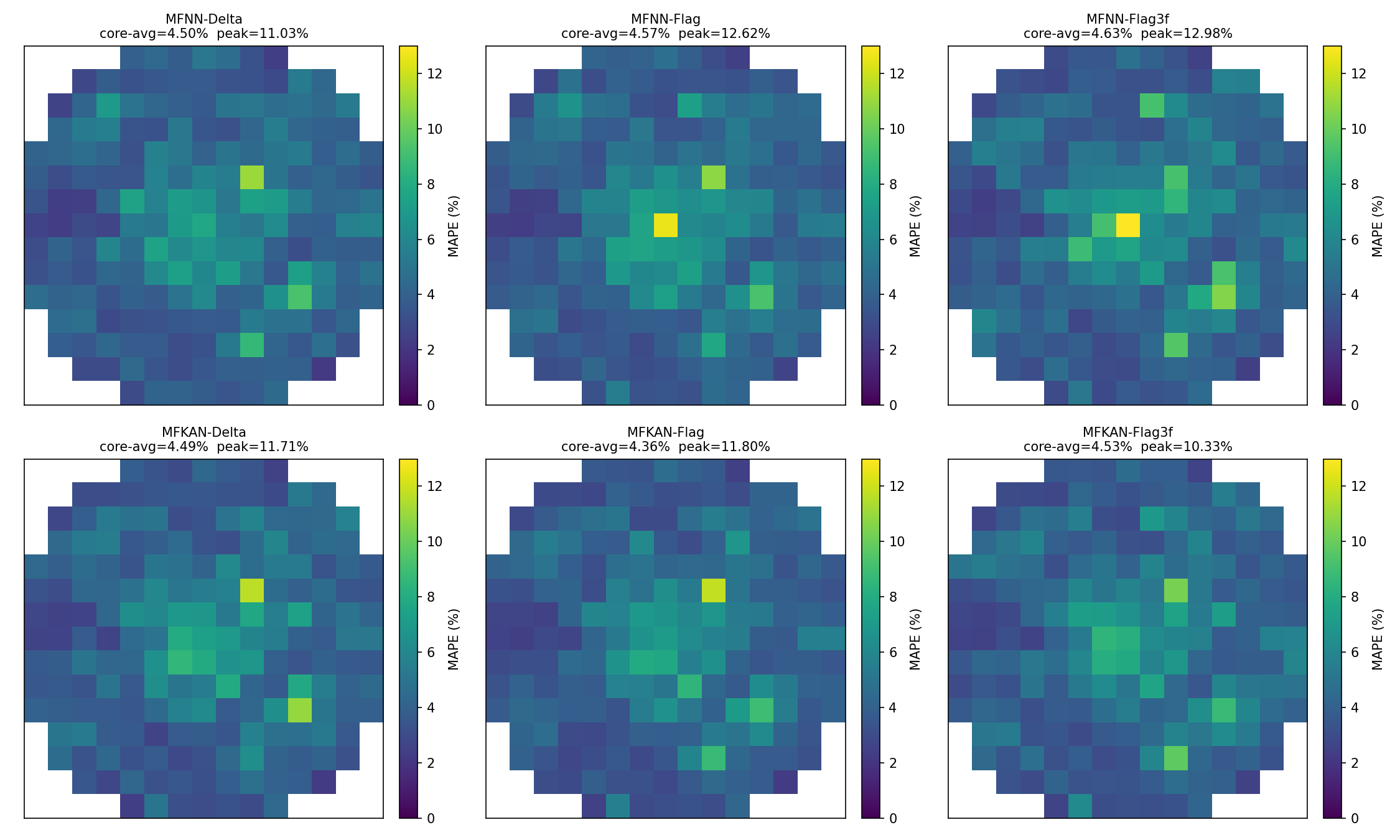}
\caption{Assembly-wise MAPE for all six multifidelity methods using 20\% of the high-fidelity data for training, with \textbf{spectral clustering applied to partition the 193 assemblies into four clusters prior to training}.}
\label{fig:real-data-mape-comparison-clustered}
\end{figure}

This section repeats the six-method MFNN versus MFKAN comparison from Section~\ref{sec:results-sf-vs-mf}, but first applies spectral clustering (Section~\ref{sec:clustering}) to partition the 193 assemblies into four clusters, with a separate network trained for each cluster. The same 20\%/80\% HF training/test split is maintained. The inputs and outputs are PCA-compressed to three components rather than the 10 used in the baseline case, reflecting the reduced dimensionality already achieved through spectral clustering. Figure~\ref{fig:real-data-mape-comparison-clustered} presents the results, showing that clustering degrades both MFNN and MFKAN performance relative to the baseline case, though MFKAN degrades noticeably less, particularly in peak MAPE -- reversing the baseline ranking where MFNN had been slightly ahead. Hyperparameters were tuned only on the baseline case (Section~\ref{sec:hyperparameter-tuning}) and reused as-is here, which likely contributes to this across-the-board drop.

\subsubsection{Most Correlated}
\label{sec:results-correlated-subset}

This experiment tests the hypothesis that multifidelity methods benefit from stronger correlations between fidelity levels, which is also supported by the results here. The analysis is restricted to the 50 most strongly correlated assemblies (Section~\ref{sec:correlated-subset-method}), rather than the full 193-assembly core. Therefore, the absolute errors are not directly comparable to those of the baseline case in Section~\ref{sec:results-sf-vs-mf}; instead, this experiment provides a relative assessment of how stronger cross-fidelity correlation affects multifidelity performance.

\begin{figure}[!ht]
\centering
\includegraphics[width=0.9\linewidth]{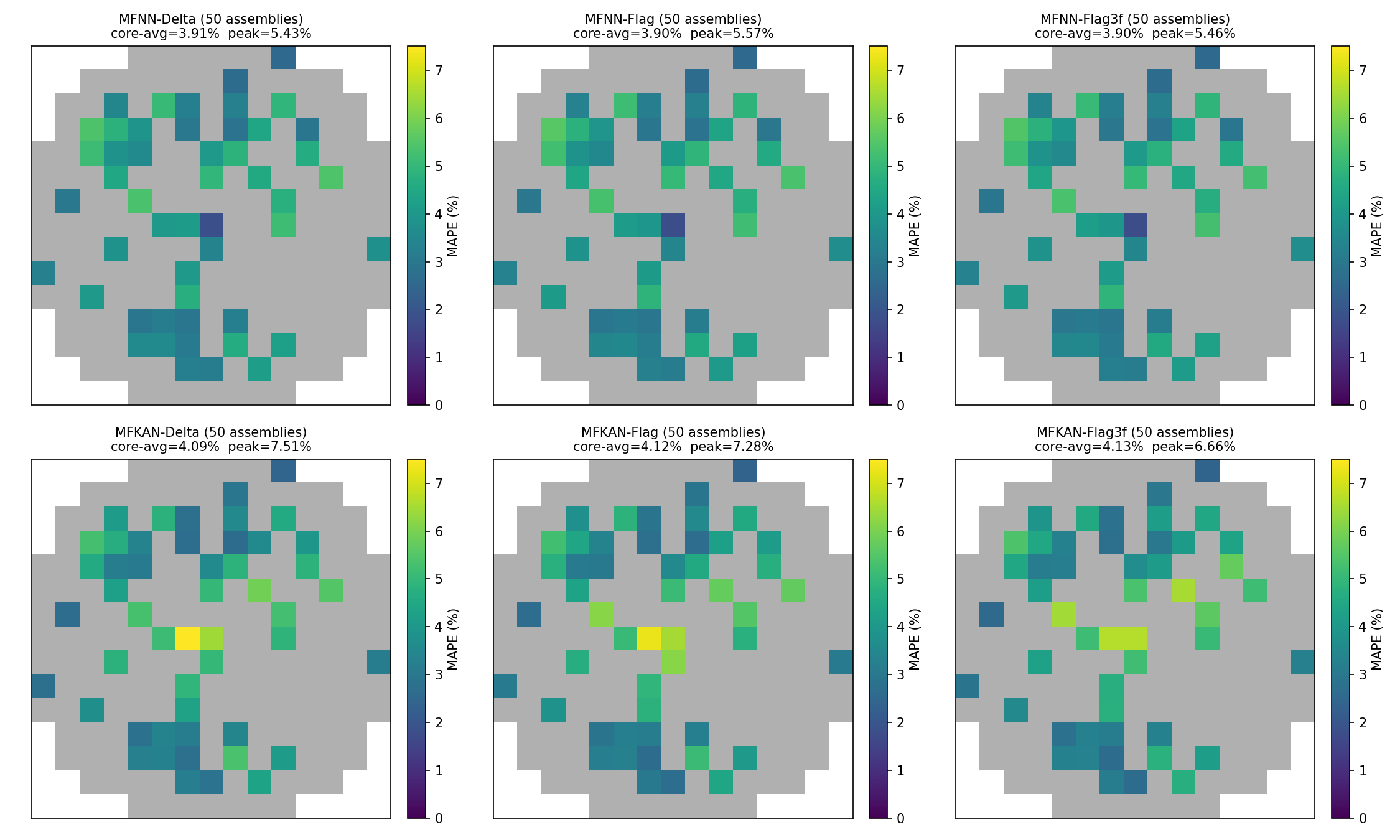}
\caption{Assembly-wise MAPE for all six multifidelity methods using 20\% of the high-fidelity data for training, evaluated on \textbf{a subset of the 50 assemblies with the highest cross-fidelity correlation}.}
\label{fig:real-data-mape-comparison-correlated}
\end{figure}

Figure~\ref{fig:real-data-mape-comparison-correlated} shows that all six methods achieve substantially lower errors for the highly correlated subset. The lowest core-average MAPE is 3.90\%, achieved by both MFNN-Flag and MFNN-Flag3f, while MFNN-Delta achieves the lowest peak MAPE at 5.43\%. Consistent with the baseline results, the MFNN methods outperform their MFKAN counterparts, only slightly for core-average MAPE but more substantially for peak MAPE. The improvement over the baseline case is most pronounced in peak MAPE for both architectures.

\subsubsection{One-Time-Step-Forward Spline vs.\ Symbolic KAN}
\label{sec:results-symbolic-vs-spline}

All of the MFKAN results discussed so far (Sections~\ref{sec:results-sf-vs-mf}--\ref{sec:results-correlated-subset}) use the spline form of the KAN. As we discussed before, one of the advantages of KAN over a standard neural network is that a trained spline can be converted into a symbolic equation, so this section compares that symbolic-regression equivalent against the spline it was extracted from, shown in Table~\ref{tab:symbolic-vs-spline}. A representative example equation, extracted from MFKAN-Flag on the baseline case, is reported in \ref{sec:appendix-symbolic-equation}.

Symbolic regression performs comparably to the spline representation across all groupings.

\input{tables/tab_symbolic_vs_spline.tex}

\subsection{TabDDPM Synthetic Data and Forecasting}
\label{sec:results-tabddpm-section}

This section covers the generative models and TabDDPM-based results. Since only one real CFD transient is available at each fidelity level, synthetic data is generated with TabDDPM to give the forecasting models more to train on, and used here to run true, compounding forecasting rather than the one-time-step-forward method in Section~\ref{sec:results-one-step-forward}. There are five parts to this section: how the synthetic data was generated and validated against the real transients (Section~\ref{sec:results-tabddpm}), hyperparameter tuning results (Section~\ref{sec:results-forecasting-hyperparameters}) ,the forecasting results using KAN and FNN trained on that synthetic data (Section~\ref{sec:results-forecasting}), the KAN spline-vs-symbolic comparison for those same forecasts (Section~\ref{sec:results-tabddpm-symbolic-vs-spline}), and an assembly-level analysis of those results (Section~\ref{sec:results-assembly-analysis}).

\subsubsection{Synthetic Data Generation and Validation}
\label{sec:results-tabddpm}

Synthetic data for each assembly was generated by running TabDDPM's generation loop, checking each new sample against every assembly's real trajectory, and adding the sample to whichever assembly's bucket it correlated most strongly with. This continued until every assembly had accumulated 1,000 synthetic realizations.

\begin{figure}[!ht]
\centering
\includegraphics[width=0.9\linewidth]{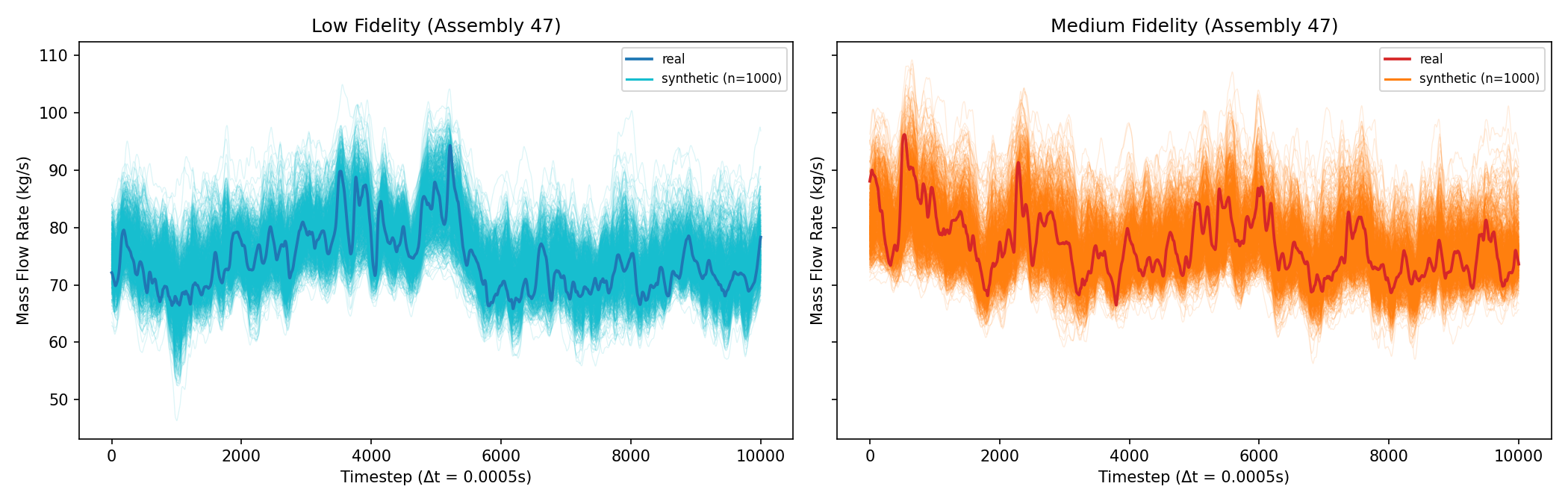}
\caption{Real and synthetic flow rate data as a function of time steps overlaid for a selected assembly (No.~47), Low-Fidelity (left) vs.\ Medium-Fidelity (right).}
\label{fig:tabddpm-lf-mf-comparison}
\end{figure}

Figure~\ref{fig:tabddpm-lf-mf-comparison} compares the synthetic and real data for a representative assembly (No.~47). Similar to the analysis in Section~\ref{sec:results-correlation}, correlation provides a more informative measure of agreement than error-based metrics such as RMSE.

\begin{figure}[!ht]
\centering
\includegraphics[width=0.9\linewidth]{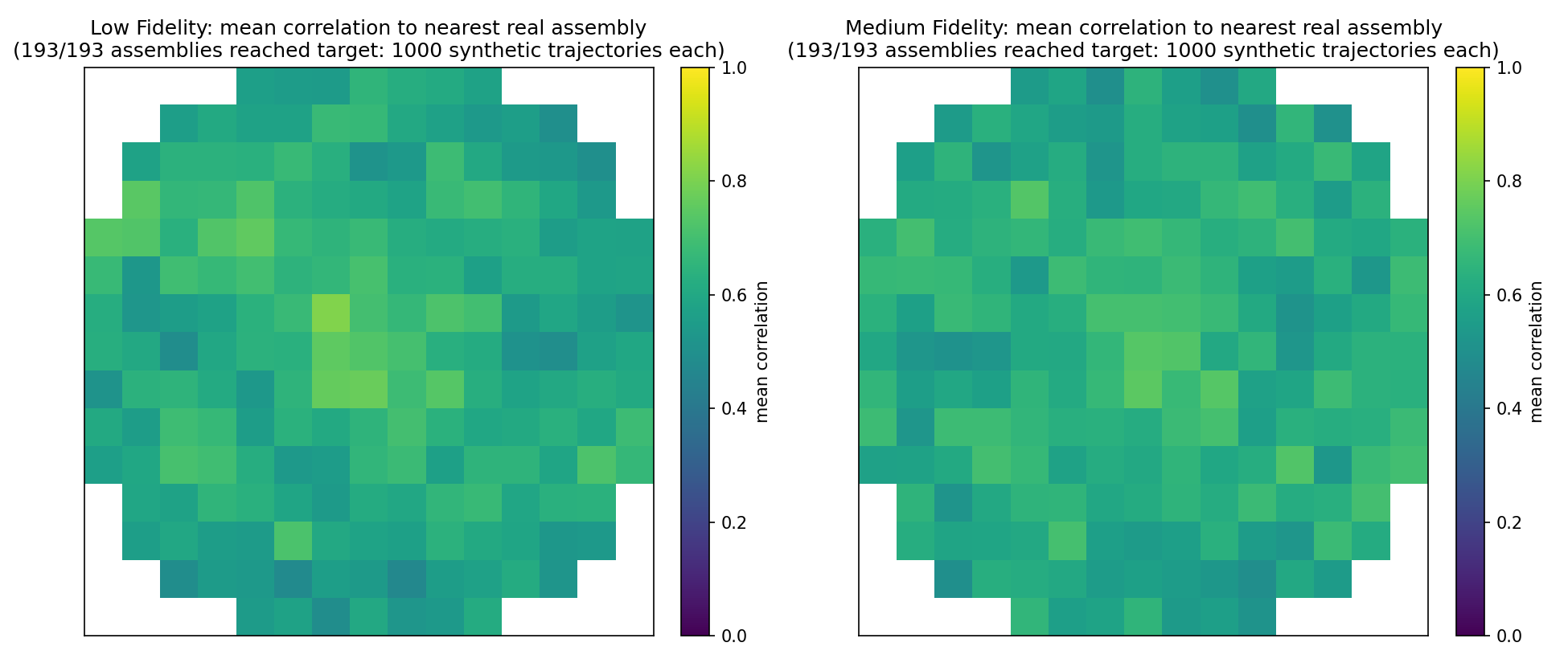}
\caption{Whole-core mean synthetic-vs-real correlation per assembly, LF and MF side by side. (Section~\ref{sec:tabddpm}).}
\label{fig:tabddpm-correlation-by-assembly}
\end{figure}

Figure~\ref{fig:tabddpm-correlation-by-assembly} presents a different measure from the cross-fidelity correlation analysis in Section~\ref{sec:results-correlation}. Here, the metric represents the mean correlation between the synthetic data generated for each assembly and its own real trajectory, indicating the usefulness of the synthetic data for assembly-level self-prediction and the cross-fidelity forecasting examined later in this section. TabDDPM generated strongly correlated synthetic samples more readily for some assemblies than others. For more difficult assemblies, the correlation threshold had to be progressively relaxed to obtain sufficient samples, resulting in millions of generated candidates before the required synthetic dataset was completed.

\subsubsection{Hyperparameter Tuning Results}
\label{sec:results-forecasting-hyperparameters}

Final tuned hyperparameters for the dedicated-fidelity KAN and FNN forecasting models (search space and tuning methodology in Section~\ref{sec:hyperparameter-tuning}) are reported in Tables~\ref{tab:hyperparameters-kan-forecasting} and \ref{tab:hyperparameters-fnn-forecasting}; the combined (LF+MF$\to$HF) models reuse the dedicated low-fidelity model's values.

\input{tables/tab_hyperparameters_kan_forecasting.tex}
\input{tables/tab_hyperparameters_fnn_forecasting.tex}

\subsubsection{Augmented Forecasting Results}
\label{sec:results-forecasting}

This section shows the results of forecasting using static methods, comparing KAN and FNN. These are no longer MFNN/MFKAN per our typical definition in Section \ref{sec:multifidelity-methods}. They nonetheless retain a multifidelity character, since the synthetic training data spans all three fidelity levels, despite each model being trained and evaluated using a single-fidelity approach. We first consider self-evaluation, where a model trained on synthetic data from a given fidelity predicts the corresponding real transient at that same fidelity.

\begin{figure}[!h]
\centering
\includegraphics[width=0.95\linewidth]{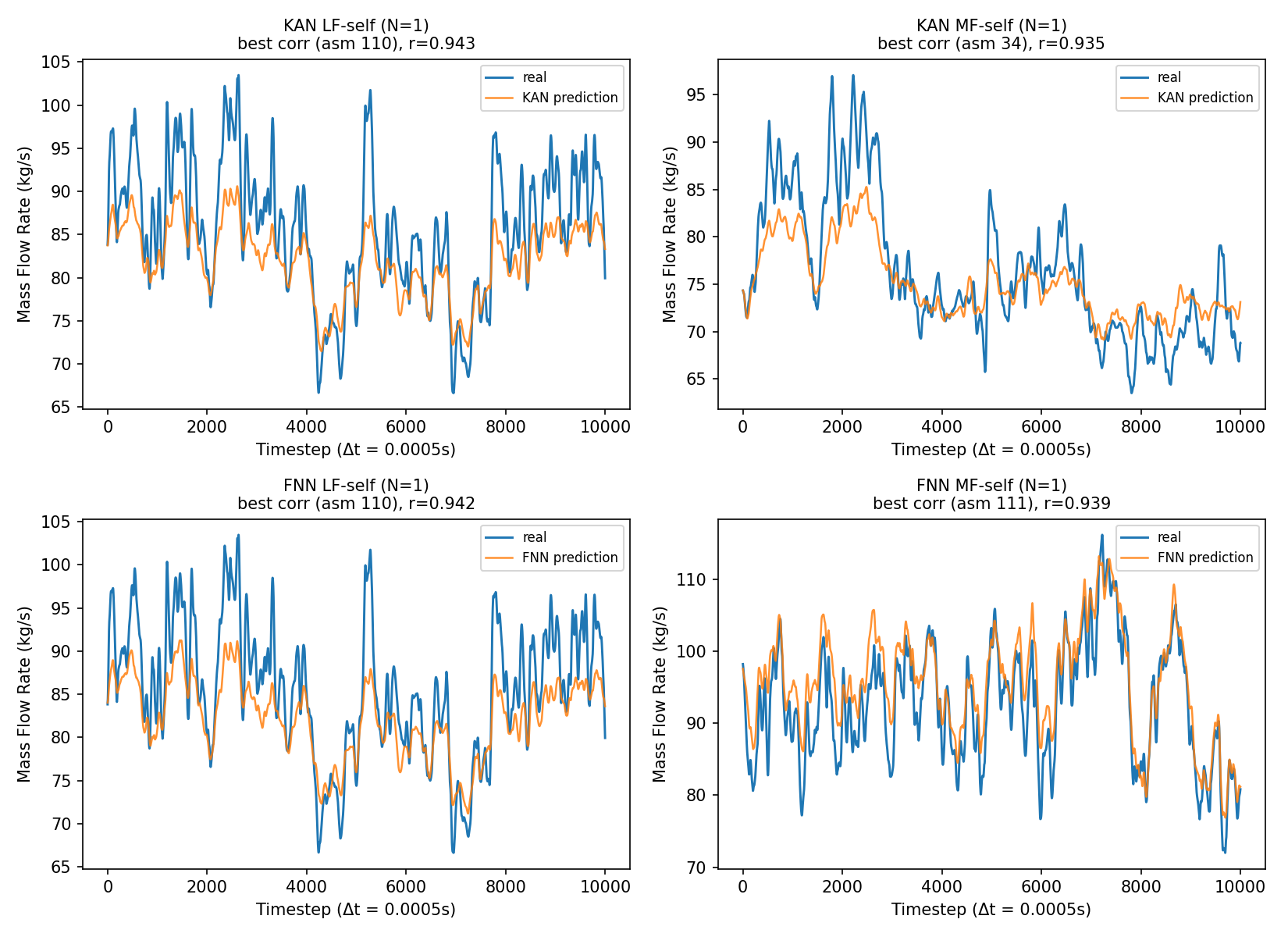}
\caption{Self-evaluation performance of KAN and FNN using synthetic data with a single initial condition ($N=1$). Rows correspond to the KAN (top) and FNN (bottom) architectures, while columns correspond to low-fidelity (LF-self, left) and medium-fidelity (MF-self, right) evaluation. Each panel compares the predicted and real mass flow rate transient for the assembly with the highest Pearson correlation ($r$), reported above the panel. The KAN model used in this figure is a spline KAN.}
\label{fig:tabddpm-selfeval-kan-fnn-n1}
\end{figure}

Figure~\ref{fig:tabddpm-selfeval-kan-fnn-n1} shows that KAN and FNN achieve comparable self-evaluation fits, with neither architecture consistently capturing the sharp peaks better than the other. This is also reflected quantitatively in Table~\ref{tab:tabddpm-fnn-vs-kan}, where the LF$\rightarrow$LF and MF$\rightarrow$MF cases denote single-fidelity self-evaluation. Using only the initial mass flow rate condition ($N=1$), KAN achieves self-evaluation correlations of 0.80--0.81 across these two cases, closely matched by FNN's 0.78--0.81.

\input{tables/tab_tabddpm_fnn_vs_kan.tex}

\begin{figure}[!h]
\centering
\includegraphics[width=\linewidth]{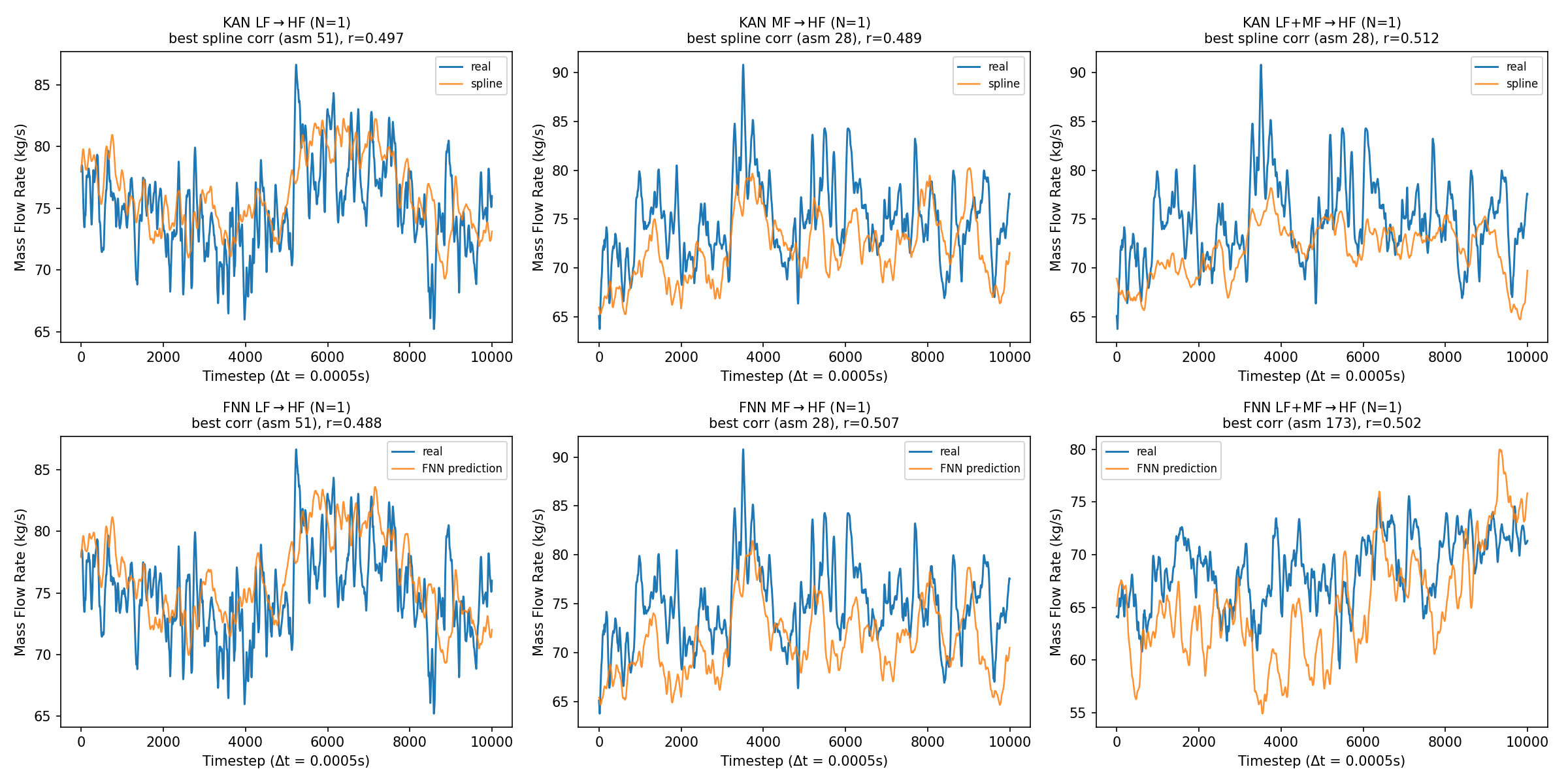}
\caption{Cross-fidelity forecasting performance of KAN and FNN using synthetic data with a single initial condition ($N=1$). Rows correspond to KAN (top) and FNN (bottom), while columns correspond to LF$\rightarrow$HF (left), MF$\rightarrow$HF (center), and combined LF+MF$\rightarrow$HF (right) forecasting. Each panel compares the predicted and real HF mass flow rate transient for the assembly with the highest Pearson correlation ($r$), reported above the panel. KAN results shown here use the learned spline representation.}
\label{fig:tabddpm-predictions-kan-fnn-n1}
\end{figure}

Figure~\ref{fig:tabddpm-predictions-kan-fnn-n1} presents the cross-fidelity forecasting case, where all synthetic data for a given LF/MF combination are used to predict the real high-fidelity transient. Each panel shows the best-performing assembly for the corresponding method and fidelity direction. Even for these best-performing cases, neither method fully captures the real transient behavior. Notably, both KAN and FNN achieve their best performance for the same assembly in two of the three directions (assembly 51 for LF$\to$HF and assembly 28 for MF$\to$HF), further motivating an examination of whether forecasting performance is related to the cross-fidelity correlations identified in Section~\ref{sec:results-correlation}. This analysis will be presented later in Section \ref{sec:results-assembly-analysis}.

Table~\ref{tab:tabddpm-fnn-vs-kan} reports the complete results, including the $N=10$ case, where the first 10 mass flow rate timesteps from the real HF simulation are provided as initial conditions rather than only the first timestep. Performance degrades at $N=10$ relative to $N=1$, likely due to the increasing input dimensionality. KAN's self-evaluation correlation decreases from 0.80--0.81 at $N=1$ to 0.73 at $N=10$, and FNN degrades similarly, from 0.78--0.81 to 0.73--0.76 -- the two architectures remain comparable in self-evaluation at both $N$. Both methods perform poorly in cross-fidelity forecasting at either $N$, with correlations of only 0.03--0.06. This behavior is consistent with the weak cross-fidelity correlations identified in Section~\ref{sec:results-correlation}.

\subsubsection{TabDDPM Spline vs.\ Symbolic KAN}
\label{sec:results-tabddpm-symbolic-vs-spline}

All KAN forecasting results in Section~\ref{sec:results-forecasting} use the learned spline representation. Following the analysis in Section~\ref{sec:results-symbolic-vs-spline}, this section compares the symbolic-regression equations extracted from the trained KANs against their original spline representations.

Table~\ref{tab:tabddpm-symbolic-vs-spline} shows that symbolic regression outperforms the spline representation across all six forecasting directions. The spline metrics reported here are recomputed by reloading each saved model, so they differ slightly (by roughly 6--8\% in RMSE) from the corresponding rows in Table~\ref{tab:tabddpm-fnn-vs-kan}, reflecting minor differences between the two evaluation pipelines rather than a different underlying model. The spline fit itself is comparatively weak in this cross-fidelity forecasting setting, with RMSE ranging from 5.70 to 6.42~kg/s. Symbolic regression reduces this error in every direction, by 0.6--9.2\%, with the largest reductions for the combined LF+MF$\to$HF direction (9.2\% at both $N=1$ and $N=10$) and for MF$\to$HF at $N=10$, where the spline error is highest (6.42 to 5.86~kg/s, 8.8\%). The lowest symbolic error is for LF+MF$\to$HF at $N=1$, where RMSE drops to 5.18~kg/s from 5.70~kg/s for the spline.

\input{tables/tab_tabddpm_symbolic_vs_spline.tex}



\subsubsection{Assembly Analysis}
\label{sec:results-assembly-analysis}

\begin{figure}[!ht]
\centering
\includegraphics[width=0.7\linewidth]{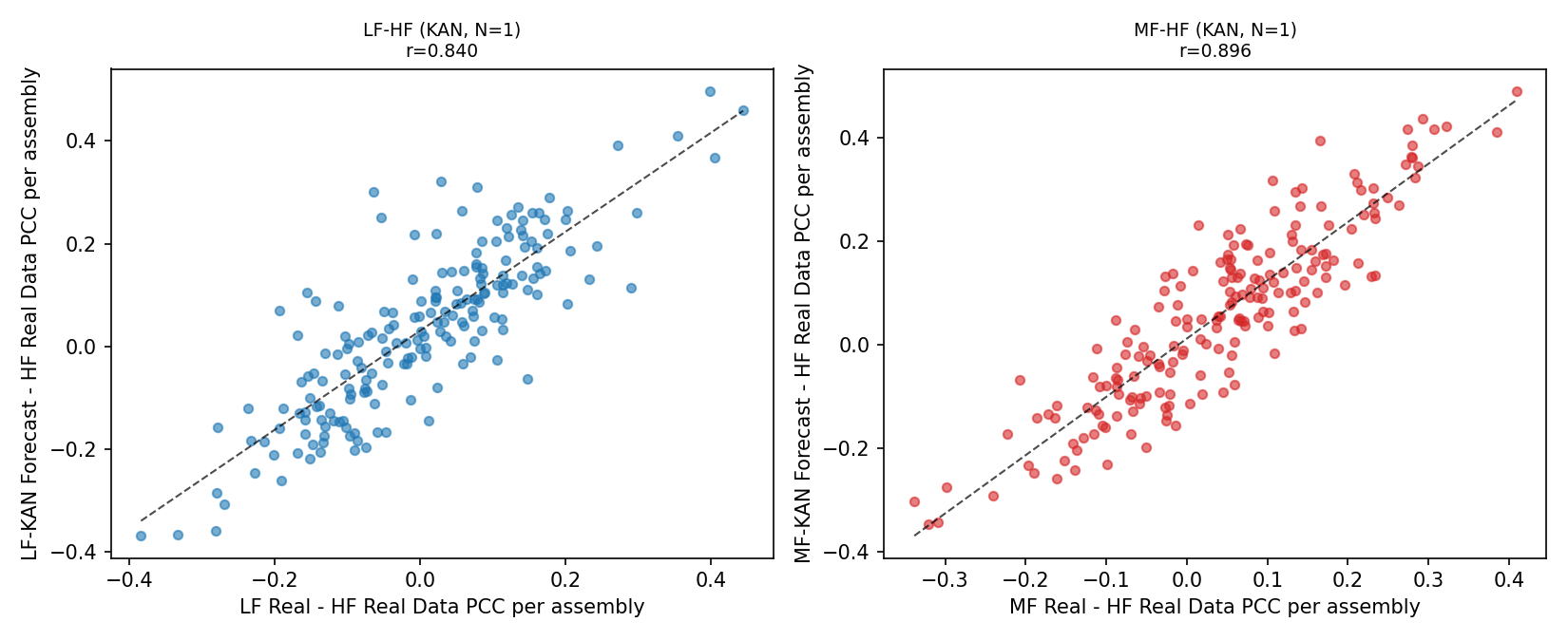}
\caption{Assembly-level relationship between real-data cross-fidelity correlation and KAN forecasting performance for $N=1$. Each point represents one fuel assembly. The horizontal axis gives the correlation between the real LF or MF transient and the real HF transient, while the vertical axis gives the correlation between the KAN-predicted and real HF transients. Results are shown for LF$\to$HF (left) and MF$\to$HF (right), with the Pearson correlation ($r$) between these two assembly-level quantities reported above each panel.}
\label{fig:tabddpm-correlation-vs-forecast-kan-n1}
\end{figure}
\begin{figure}[!ht]
\centering
\includegraphics[width=0.7\linewidth]{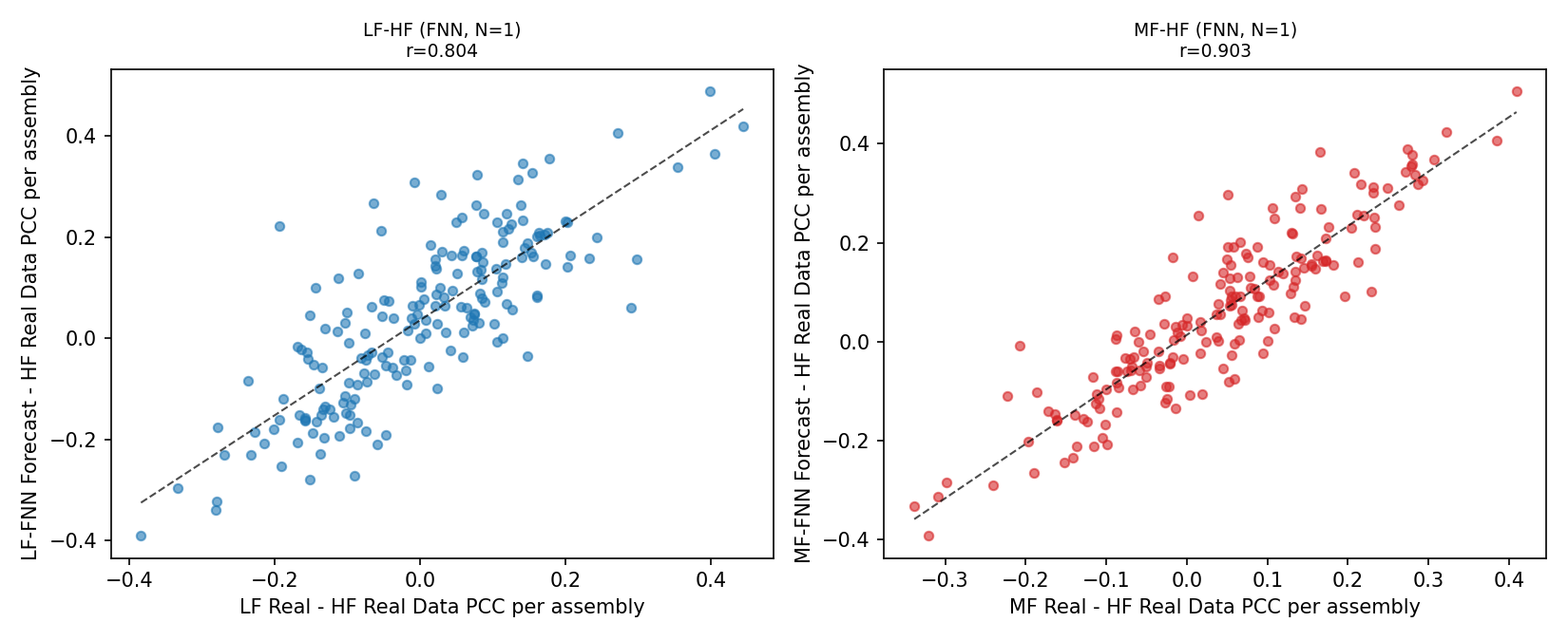}
\caption{Assembly-level relationship between real-data cross-fidelity correlation and FNN forecasting performance for $N=1$. Each point represents one fuel assembly. The horizontal axis gives the correlation between the real LF or MF transient and the real HF transient, while the vertical axis gives the correlation between the FNN-predicted and real HF transients. Results are shown for LF$\to$HF (left) and MF$\to$HF (right), with the Pearson correlation ($r$) between these two assembly-level quantities reported above each panel.}
\label{fig:tabddpm-correlation-vs-forecast-fnn-n1}
\end{figure}

In Section~\ref{sec:results-forecasting}, both KAN and FNN achieved their best forecast performance for the same assembly in two of the three directions: assembly 51 for LF$\to$HF and assembly 28 for MF$\to$HF. This observation motivates an assembly-level analysis of whether forecast performance is related to the underlying correlation between fidelity levels. Figures~\ref{fig:tabddpm-correlation-vs-forecast-kan-n1} and \ref{fig:tabddpm-correlation-vs-forecast-fnn-n1} compare these quantities for KAN and FNN, respectively. Each point represents one fuel assembly: the horizontal axis gives the real-data cross-fidelity correlation between the LF or MF transient and the corresponding HF transient, while the vertical axis gives the correlation between the model-predicted and real HF transients for that assembly. \textit{Ideally, stronger real-data cross-fidelity correlation should correspond to forecast correlations approaching unity, producing a strong positive relationship across assemblies}. Such a relationship would indicate that assemblies with greater similarity between fidelity levels provide more transferable information for HF forecasting.

For KAN, this relationship is strong, with $r=0.840$ for LF$\to$HF and $r=0.896$ for MF$\to$HF. The relationship is similarly strong for FNN, with $r=0.804$ for LF$\to$HF and $r=0.903$ for MF$\to$HF. Thus, for both architectures, assemblies with stronger real-data cross-fidelity correlation generally exhibit higher forecast correlation, indicating that both KAN and FNN predict the HF transient more effectively when stronger transferable structure exists between fidelity levels. This provides an assembly-level explanation for the challenging full-core forecasting results in Section~\ref{sec:results-forecasting}: when the underlying fidelities are weakly correlated, there is limited information from which to infer the HF transient, regardless of architecture.

\section{Discussion}
\label{sec:discussion}

Predicting the transient flow field considered in this work presents a particularly challenging multifidelity machine learning problem: cross-fidelity correlations are weak across most of the core (Section~\ref{sec:results-correlation}), the underlying transients are highly oscillatory and spatially heterogeneous, and only a single transient is available for training at each fidelity level. The central question motivating this work is whether multifidelity learning can reduce the amount of expensive high-fidelity data required to predict this transient, evaluated here using both MFNN and MFKAN, with MFKAN additionally offering interpretability through closed-form symbolic equations unavailable with MFNN if it can match MFNN's accuracy. Neither architecture performs particularly well in absolute terms, especially for TabDDPM-augmented forecasting (Section~\ref{sec:results-forecasting}). MFKAN nonetheless remains competitive with MFNN throughout. This holds for the baseline, clustered, and correlated-subset one-time-step-forward cases (Section~\ref{sec:results-one-step-forward}), as well as for TabDDPM self-evaluation (correlations of roughly 0.73--0.81 for both) and true cross-fidelity forecasting.

Cross-fidelity correlation emerges as the main driver of achievable accuracy, independent of architecture. The correlated-assembly subset (Section~\ref{sec:results-correlated-subset}) improves core-average and peak MAPE substantially over the full-core baseline for both MFNN and MFKAN, so the benefit of multifidelity learning in this problem is assembly-dependent rather than uniform across the core, concentrated in assemblies where the fidelity levels are already well correlated. These correlated-subset results, however, have limited practical applicability on their own, since they consider only the 50 assemblies already identified as having strong cross-fidelity correlations rather than enabling full-core prediction. The assembly-level analysis in the TabDDPM setting (Section~\ref{sec:results-assembly-analysis}) supports this same correlation-driven mechanism directly: both KAN's and FNN's per-assembly forecast correlation track real cross-fidelity correlation comparably closely ($r=0.840$--$0.896$ for KAN versus $r=0.804$--$0.903$ for FNN), indicating both architectures are similarly sensitive to the physical cross-fidelity structure actually present in the data, rather than this being a KAN-specific advantage.

Clustering degrades one-time-step-forward accuracy for both MFNN and MFKAN relative to the baseline case, though MFKAN noticeably less so, reversing the baseline ranking (Section~\ref{sec:results-clustering}). This degradation is likely attributable, at least in part, to hyperparameters being tuned only on the baseline case and reused as-is for the clustered case rather than with a clustering-specific search; it may also suggest that spectral clustering is not a particularly effective preprocessing step for this problem.

All one-time-step-forward results use a fixed 20\% high-fidelity training split, where the complete low- and medium-fidelity transients and the first 20\% of the high-fidelity transient are available for training (Section~\ref{sec:results-one-step-forward}). Additional real high-fidelity data would likely improve prediction over the remaining unseen portion. However, testing that directly would require more real HF CFD data, and that same scarcity is what the TabDDPM-based data augmentation in this work is designed to address instead (Section~\ref{sec:results-tabddpm-section}). Repeating this analysis across different real HF training splits was therefore not pursued.

The TabDDPM-augmented forecasts are considerably weaker than the one-time-step-forward predictions, with cross-fidelity correlations of only 0.03--0.06 (Table~\ref{tab:tabddpm-fnn-vs-kan}). Three factors likely explain this gap. First, the forecasting task is harder: unlike the one-time-step-forward models, the forecasting models see no real HF data during training and propagate their own predictions, so errors compound. Second, synthetic augmentation adds data quantity but not cross-fidelity information: the LF and MF samples reproduce their own fidelity well (self-evaluation correlations of 0.73--0.81) but carry no more information about HF than the real data, so forecast accuracy follows the underlying cross-fidelity correlation (Section~\ref{sec:results-assembly-analysis}). Third, all samples derive from a single real transient per fidelity level, which limits their diversity. Generative augmentation of lower-fidelity data alone is therefore not a substitute for real HF information when fidelities are weakly correlated; generative models conditioned on LF or MF inputs to produce HF-like samples are a promising alternative.

KAN's interpretability advantage, through the symbolic regression it enables (Section~\ref{sec:kan}), persists across both experimental settings. For one-time-step-forward prediction, the extracted symbolic equations match the spline representation's accuracy rather than exceeding it (Section~\ref{sec:results-symbolic-vs-spline}); for the TabDDPM-augmented forecasting experiments, symbolic regression instead improves on the spline representation across all six forecasting directions, most substantially for the combined LF+MF$\to$HF direction and for MF$\to$HF at $N=10$, where RMSE drops from 6.42~kg/s (spline) to 5.86~kg/s (symbolic) (Section~\ref{sec:results-tabddpm-symbolic-vs-spline}). Even this improved fit is low in absolute terms, reflecting how difficult the underlying forecasting task is rather than a shortcoming specific to symbolic regression. Taken together, these results indicate that MFKAN's interpretability does not come at an additional accuracy cost beyond what FNN and the spline representation already pay in this difficult forecasting setting. This interpretability is also not limited by the PCA compression used throughout this work (Section~\ref{sec:pca}). The equations are reported in PCA-component space (\ref{sec:appendix-symbolic-equation}), and recovering the physical mass flow rate from them only requires applying the PCA inverse transform, a quick linear operation also provided in the accompanying repository.


Some limitations of this study should be noted. Only a single transient is available at each fidelity level, rather than many independent realizations, which limits how far these results can be expected to generalize (Section~\ref{sec:fidelity}). MFKAN also showed some run-to-run variance, particularly for MFKAN-Delta, and much less so for MFKAN-Flag and MFKAN-Flag3f (Table~\ref{tab:reproducibility}); consistent with this, MFKAN-Flag generally achieved slightly better accuracy than MFKAN-Delta across the one-time-step-forward experiments, though typically by a modest margin. The TabDDPM synthetic-data validation reported in Section~\ref{sec:results-tabddpm} is limited to correlation-based comparisons; a fuller statistical or physical validation of the synthetic transients has not yet been explored and may become a subject for future work.

A natural extension of this work is to move beyond the static prediction methods considered here toward a fully autoregressive, time-dependent architecture, where predictions are recursively used as inputs for subsequent timesteps rather than generated independently. Time-dependent KAN architectures have already been proposed, such as TKAN \citep{genet2025tkan}, which incorporates recurrent memory into the KAN framework for multi-step time-series forecasting; however, to our knowledge, such temporal KAN architectures have not yet been combined with multifidelity learning. These architectures were intentionally not considered in the present study because introducing recurrent dependencies would likely increase the complexity of the resulting symbolic expressions and diminish the interpretability that motivates the use of KAN. A future extension could therefore shift the emphasis from symbolic interpretability toward improved long-term forecasting accuracy, investigating whether temporal multifidelity KAN architectures can provide more accurate multi-step predictions despite the associated increase in model complexity.

\section{Conclusions}
\label{sec:conclusions}

This work evaluated whether multifidelity surrogate modeling, using three CFD fidelity levels, can reduce the amount of expensive high-fidelity data required to predict the spatiotemporal mass flow rate field across all 193 fuel assemblies of a pressurized water reactor core. Multifidelity feedforward neural networks (MFNNs) and multifidelity Kolmogorov-Arnold Networks (MFKANs) were compared for this task, with MFKAN offering interpretability as an additional benefit contingent on matching MFNN's predictive accuracy. The problem was examined through both one-time-step-forward prediction and TabDDPM-augmented cross-fidelity forecasting under challenging conditions characterized by weak cross-fidelity correlations, highly oscillatory and spatially heterogeneous transients, and only one available CFD transient per fidelity level. For one-time-step-forward prediction, MFKAN remained competitive with MFNN across the baseline, clustered, and highly correlated assembly cases, while the strong improvement observed for the correlated subset demonstrated the importance of cross-fidelity similarity for effective multifidelity learning. In the synthetic-data experiments, KAN and FNN achieved comparable self-fidelity forecasting accuracy (correlations of roughly 0.73--0.81 for both), and both architectures struggled with true cross-fidelity forecasting. Assembly-level analysis further showed that both KAN's and FNN's forecasting performance was strongly associated with the underlying cross-fidelity correlation, to a comparable degree for both architectures. Beyond predictive performance, MFKAN provides an important advantage over MFNN through its ability to extract closed-form symbolic representations. These equations matched or exceeded the accuracy of their spline counterparts in both the one-time-step-forward and TabDDPM-augmented settings, though absolute accuracy in the latter setting remained modest for both representations, with symbolic RMSE of 5.18--5.86~kg/s across the six forecasting directions. Overall, these results show that KAN-based multifidelity modeling achieves predictive accuracy comparable to conventional neural-network approaches, while additionally offering interpretability unavailable from MFNN, even though both architectures' absolute accuracy was often modest in the more difficult forecasting settings. Cross-fidelity correlation and the limited availability of transient data, rather than any inherent limitation of either architecture, emerge as the primary constraints on this full-core multifidelity forecasting problem.

\section*{Data Availability}
\label{sec:data_avail}

The authors have all the data and codes to reproduce all the results in this work currently in a private GitHub repository. To ensure the confidentiality of this research, the authors will make this repository public during an advanced stage of the review process, which will be listed under our research group's public Github page: \url{https://github.com/aims-umich}

\section*{Acknowledgment}
This research is sponsored by the Department of Energy Office of Nuclear Energy through the Nuclear Energy University Programs (Award number: DE-NE0009506). The authors would like to thank Nataly Panczyk and Hyungjun Kim for their assistance with initial code implementation and for reviewing and editing this manuscript.

\section*{AI Usage Statement}

The authors used Claude and Claude Code (Anthropic) as assistive tools in this work for manuscript writing, including language editing, grammar correction, and clarity improvement; literature assistance, including citation verification against original sources and search-term generation; coding, including code generation, debugging, and documentation for data processing, figure and table generation, and paper-asset pipeline scripts; data analysis, including computing and verifying summary statistics reported in the Results section; and visualization, including plotting scripts and figure generation. All cited references were independently located and verified, all AI-assisted code was reviewed and tested, numerical results were independently checked, and the underlying data, scientific content, interpretations, arguments, and conclusions were developed and verified by the authors. \textit{All AI-generated or AI-modified material was reviewed and, where applicable, independently tested or validated, and the authors take full responsibility for the final content and conclusions}.

\section*{CRediT Author Statement}

\begin{itemize}

    \item \textbf{Meredith Eaheart}: Conceptualization, Methodology, Software, Formal Analysis, Visualization, Investigation, Data Curation, Writing - Original Draft.
    \item \textbf{Majdi I. Radaideh}: Conceptualization, Methodology, Software, Validation, Data Curation, Funding Acquisition, Supervision, Project Administration, Writing - Review and Edit.

\end{itemize}


\bibliographystyle{elsarticle-num}
\setlength{\bibsep}{0pt plus 0.3ex}
{\small
\bibliography{master_reference}}

\clearpage

\appendix
\section{Example Symbolic Equation}
\label{sec:appendix-symbolic-equation}

Every trained KAN method in this work produces its own closed-form symbolic equation (Section~\ref{sec:kan}), and reporting every one explicitly is not practical given the number of method, assembly-grouping, and fidelity combinations considered. As a representative example, this appendix reports the equation extracted from MFKAN-Flag trained on the baseline, full-core case (Section~\ref{sec:results-sf-vs-mf}). The full set of extracted equations for every method and experiment, along with the PCA components, scalers, and the exact transform needed to evaluate them on raw physical inputs, is available in the group GitHub repository (Data Availability).

The variables $x_1, \ldots, x_{10}$ denote the 10 PCA-compressed input components (Section~\ref{sec:pca}); \texttt{Y\_PC1} through \texttt{Y\_PC10} denote the 10 PCA-compressed output components. Recovering the physical mass flow rate at each of the 193 assemblies requires inverting the PCA transform, detailed in the repository alongside each equation.

\begin{lstlisting}[breaklines=true, basicstyle=\small\ttfamily, frame=single]
Y_PC1 = -9.2741*exp(-0.0946*x_1 - 2.1883*cos(0.2572*x_3 + 0.7199))/(3.4453*x_4 + 7.3027)**0.805 + 0.3629*cos(0.7525*x_9 + 8.1702) - 13.4042*cos(0.0952*x_1 + 0.0358*x_9 - 0.2571*log(7.2523 - 3.9202*x_8) + 0.0601*tanh(0.9209*x_3 - 0.9073) + 3.2689) - 10.7336
Y_PC2 = 0.9787*x_2 + 0.0121
Y_PC3 = 2.9528*sin(0.3538*x_3 + 6.5209) - 0.6739
Y_PC4 = -1.3144*sqrt(4.4384 - 1.9573*x_3) + 3.8869*sin(0.0433*x_1 + 0.3688*log(3.4453*x_4 + 7.3027) + 1.0027*cos(0.2572*x_3 + 0.7199) - 7.8439) + 3.2508
Y_PC5 = -6.6252*cos(0.1643*x_5 - 4.4358) - 1.8247
Y_PC6 = -3.7527*sin(-0.2771*x_6 + 0.0154*sin(1.0028*x_1 - 7.7918) + 0.0329*sin(0.9741*x_3 - 3.2123) + 5.9715) + 0.0616*sin(0.1934*x_1 + 0.0727*x_9 - 0.522*log(7.2523 - 3.9202*x_8) + 0.122*tanh(0.9209*x_3 - 0.9073) - 4.0911) + 0.1294*cos(0.8587*x_3 + 2.0466) - 1.1524
Y_PC7 = 1.0048*sin(0.139*x_1 + 1.1828*log(3.4453*x_4 + 7.3027) + 3.2154*cos(0.2572*x_3 + 0.7199) - 5.548) - 0.6751*cos(0.8276*x_3 + 8.0656) + 1.978*cos(0.2054*x_5 - 2.47) - 1.7418*cos(0.1821*x_1 + 0.0685*x_9 - 0.4916*log(7.2523 - 3.9202*x_8) + 0.1149*tanh(0.9209*x_3 - 0.9073) + 4.8292) + 0.2833
Y_PC8 = -4.5294*sin(0.2598*x_6 - 0.0144*sin(1.0028*x_1 - 7.7918) - 0.0309*sin(0.9741*x_3 - 3.2123) + 4.6926) + 2.1318*sin(0.2669*x_1 + 0.1004*x_9 - 0.7205*log(7.2523 - 3.9202*x_8) + 0.1683*tanh(0.9209*x_3 - 0.9073) - 5.4731) + 0.5057*cos(0.7073*x_9 + 8.8228) - 3.6405
Y_PC9 = -2.218*sin(0.4388*x_9 + 9.4949) - 0.1349
Y_PC10 = 0.9887*x_10 - 0.0022
\end{lstlisting}

\end{document}

%% file: tables/tab_hyperparameters_search.tex
\begin{table}[H]
\centering
\caption{Hyperparameter search space for all four tuned model families. MFKAN/MFNN (baseline) ran up to 100 Optuna TPE trials per method to completion. Model group KAN covers MFKAN and KAN forecasting; FNN covers MFNN and FNN forecasting.}
\label{tab:hyperparameters-search}
\begin{tabular}{p{1.6cm}p{3.6cm}p{5.8cm}}
\toprule
Model & Hyperparameter & Search Space \\
\midrule
KAN & Depth (hidden layers) & $\{1, 2\}$ \\
 & Hidden width (MFKAN) & 8 (fixed, not tuned) \\
 & Hidden width (KAN forecast.) & 32 (fixed, not tuned) \\
 & Grid size & $\{3,4,5,6,7,8,9,10\}$ \\
 & Spline order $k$ & $\{2,3,4,5,6,7,8\}$ \\
 & Steps & $\{25,50,75,100,125,150,200,250\}$ \\
 & $\lambda$ & $[0, 0.001]$ \\
 & $\lambda_{entropy}$ & $[0, 10]$ \\
 & Learning rate & $[10^{-4}, 10^{-1}]$ (log) \\
 & Reg. metric & \texttt{edge\_forward\_spline\_n}, \texttt{edge\_forward\_sum}, \texttt{edge\_forward\_spline\_u}, \texttt{edge\_backward}, \texttt{node\_backward} \\
FNN & Hidden layers (count) & $\{2, 3, 4\}$ \\
 & Nodes per layer & $\{8,16,32,64,128,256\}$ \\
 & Learning rate & $[10^{-5}, 10^{-1}]$ (log) \\
 & Epochs (MFNN) & $\{250,500,1000,1500,2000,3000,4000,5000\}$ \\
 & Epochs (FNN forecast.) & $\{100,200,300,400,500,650,800,1000\}$ \\
 & Weight decay & $[10^{-6}, 10^{-2}]$ (log) \\
\bottomrule
\end{tabular}
\end{table}

%% file: tables/tab_hyperparameters_mfkan.tex
\begin{table}[H]
\centering
\caption{MFKAN baseline hyperparameters (tuned).}
\label{tab:hyperparameters-mfkan}
\begin{tabular}{lccccccccc}
\toprule
Method & Depth & Hidden & Grid & k & Steps & Learning Rate & $\lambda$ & $\lambda_{entropy}$ & Reg. Metric \\
\midrule
Delta & 1 & 8 & 4 & 8 & 75 & 0.011 & 8.568e-04 & 6.7996 & \texttt{edge\_backward} \\
Flag & 1 & 8 & 3 & 8 & 250 & 0.0079 & 7.557e-04 & 6.7701 & \texttt{node\_backward} \\
Flag3f & 1 & 8 & 3 & 7 & 75 & 0.092 & 5.761e-04 & 4.0174 & \texttt{edge\_forward\_spline\_u} \\
\bottomrule
\end{tabular}
\end{table}

%% file: tables/tab_hyperparameters_mfnn.tex
\begin{table}[H]
\centering
\caption{MFNN baseline hyperparameters (tuned).}
\label{tab:hyperparameters-mfnn}
\begin{tabular}{lcccc}
\toprule
Method & Hidden & Learning Rate & Epochs & Weight Decay \\
\midrule
MFNN-Delta & [128, 128] & 0.0026 & 4000 & 1.059e-06 \\
MFNN-Flag & [16, 16, 16] & 0.0093 & 2000 & 3.638e-04 \\
MFNN-Flag3f & [16, 16, 16] & 0.016 & 3000 & 3.883e-05 \\
\bottomrule
\end{tabular}
\end{table}

%% file: tables/tab_reproducibility.tex
\begin{table}[h]
\centering
\caption{Reproducibility across repeated runs (mean $\pm$ std).}
\label{tab:reproducibility}
\begin{tabular}{lccccc}
\toprule
Method & N Repeats & RMSE Mean & RMSE Std & MAE Mean & MAE Std \\
\midrule
MFKAN-Delta & 5 & 4.3160 & 0.0195 & 3.2850 & 0.0145 \\
MFKAN-Flag & 5 & 4.2662 & 0.0014 & 3.2583 & 0.0018 \\
MFKAN-Flag3f & 5 & 4.2674 & 0.0026 & 3.2478 & 0.0013 \\
MFNN-Delta & 5 & 4.1733 & 0.0000 & 3.2162 & 0.0000 \\
MFNN-Flag & 5 & 4.1673 & 0.0000 & 3.2130 & 0.0000 \\
MFNN-Flag3f & 5 & 4.1672 & 0.0000 & 3.2130 & 0.0000 \\
\bottomrule
\end{tabular}
\end{table}

%% file: tables/tab_symbolic_vs_spline.tex
\begin{table}[h]
\centering
\caption{RMSE of symbolic-regression and spline representations for the MFKAN methods across the baseline, clustered, and correlated-subset cases.}
\label{tab:symbolic-vs-spline}
\begin{tabular}{lcc}
\toprule
Method & RMSE Spline & RMSE Symbolic \\
\midrule
\multicolumn{3}{l}{\textit{Plain}} \\
MFKAN-Flag3f & 4.2653 & 4.2707 \\
MFKAN-Flag & 4.2671 & 4.2689 \\
MFKAN-Delta & 4.3607 & 4.3484 \\
\midrule
\multicolumn{3}{l}{\textit{Clustered (4 binary clusters)}} \\
MFKAN-Flag3f, cluster 0 & 4.6772 & 4.5294 \\
MFKAN-Flag3f, cluster 1 & 4.3026 & 4.1853 \\
MFKAN-Flag3f, cluster 2 & 4.2055 & 4.0480 \\
MFKAN-Flag3f, cluster 3 & 5.3492 & 5.2631 \\
MFKAN-Flag, cluster 0 & 4.5131 & 4.4752 \\
MFKAN-Flag, cluster 1 & 4.1843 & 4.1969 \\
MFKAN-Flag, cluster 2 & 3.8520 & 3.7984 \\
MFKAN-Flag, cluster 3 & 5.3098 & 5.3316 \\
MFKAN-Delta, cluster 0 & 4.7328 & 4.6758 \\
MFKAN-Delta, cluster 1 & 4.3067 & 4.2487 \\
MFKAN-Delta, cluster 2 & 4.0471 & 3.8920 \\
MFKAN-Delta, cluster 3 & 5.2682 & 5.2564 \\
\midrule
\multicolumn{3}{l}{\textit{Correlated Subset (top 50)}} \\
MFKAN-Flag3f & 4.1960 & 4.2504 \\
MFKAN-Flag & 4.1980 & 4.1986 \\
MFKAN-Delta & 4.1503 & 4.1599 \\
\bottomrule
\end{tabular}
\end{table}

%% file: tables/tab_hyperparameters_kan_forecasting.tex
\begin{table}[H]
\centering
\caption{KAN forecasting hyperparameters (tuned; combined reuses LF's values -- see apply\_forecasting\_lf\_mf.py).}
\label{tab:hyperparameters-kan-forecasting}
\begin{tabular}{lccccccccc}
\toprule
Variant & Depth & Hidden & Grid & k & Steps & $\lambda$ & $\lambda_{entropy}$ & Learning Rate & Reg. Metric \\
\midrule
LF & 1 & 32 & 10 & 8 & 250 & 6.713e-06 & 0.1175 & 0.0116 & \texttt{edge\_forward\_spline\_n} \\
MF & 1 & 32 & 8 & 4 & 250 & 3.622e-05 & 4.0001 & 0.0299 & \texttt{edge\_forward\_sum} \\
COMBINED & 1 & 32 & 10 & 8 & 250 & 6.713e-06 & 0.1175 & 0.0116 & \texttt{edge\_forward\_spline\_n} \\
\bottomrule
\end{tabular}
\end{table}

%% file: tables/tab_hyperparameters_fnn_forecasting.tex
\begin{table}[H]
\centering
\caption{FNN forecasting hyperparameters (tuned; combined reuses LF's values -- see apply\_forecasting\_lf\_mf.py).}
\label{tab:hyperparameters-fnn-forecasting}
\begin{tabular}{lccccc}
\toprule
Variant & Depth & Hidden & Learning Rate & Epochs & Weight Decay \\
\midrule
LF & 4 & 128 & 0.0078 & 500 & 4.092e-06 \\
MF & 4 & 256 & 0.005 & 1000 & 1.130e-06 \\
COMBINED & 4 & 128 & 0.0078 & 500 & 4.092e-06 \\
\bottomrule
\end{tabular}
\end{table}

%% file: tables/tab_tabddpm_fnn_vs_kan.tex
\begin{table}[h]
\centering
\caption{TabDDPM cross-fidelity forecast: KAN vs.\ FNN.}
\label{tab:tabddpm-fnn-vs-kan}
\begin{tabular}{lccccccc}
\toprule
 & & \multicolumn{3}{c}{KAN} & \multicolumn{3}{c}{FNN} \\
\cmidrule(lr){3-5} \cmidrule(lr){6-8}
Direction & N & Correlation & RMSE & MAE & Correlation & RMSE & MAE \\
\midrule
LF $\to$ LF & 1 & 0.8030 & 3.0319 & 2.4323 & 0.7763 & 3.0763 & 2.4587 \\
MF $\to$ MF & 1 & 0.8059 & 3.2755 & 2.6714 & 0.8120 & 3.1850 & 2.5877 \\
LF $\to$ HF & 1 & 0.0336 & 5.5506 & 4.5281 & 0.0384 & 5.6863 & 4.6463 \\
MF $\to$ HF & 1 & 0.0582 & 5.4903 & 4.4897 & 0.0592 & 5.5980 & 4.5668 \\
LF+MF $\to$ HF & 1 & 0.0562 & 5.3724 & 4.3844 & 0.0370 & 7.2223 & 5.8852 \\
LF $\to$ LF & 10 & 0.7327 & 3.4880 & 2.8397 & 0.7567 & 3.1896 & 2.5453 \\
MF $\to$ MF & 10 & 0.7310 & 3.8932 & 3.2192 & 0.7289 & 3.7422 & 3.0649 \\
LF $\to$ HF & 10 & 0.0345 & 5.6584 & 4.6258 & 0.0405 & 5.6006 & 4.5745 \\
MF $\to$ HF & 10 & 0.0554 & 5.9511 & 4.8955 & 0.0591 & 5.9294 & 4.8780 \\
LF+MF $\to$ HF & 10 & 0.0573 & 5.5368 & 4.5427 & 0.0553 & 8.5356 & 7.0185 \\
\bottomrule
\end{tabular}
\end{table}

%% file: tables/tab_tabddpm_symbolic_vs_spline.tex
\begin{table}[h]
\centering
\caption{RMSE of symbolic-regression and spline representations for the TabDDPM-augmented cross-fidelity KAN forecasts.}
\label{tab:tabddpm-symbolic-vs-spline}
\begin{tabular}{lcc}
\toprule
Direction & RMSE Spline & RMSE Symbolic \\
\midrule
LF $\to$ HF, $N=1$ & 5.8893 & 5.8512 \\
MF $\to$ HF, $N=1$ & 5.8348 & 5.5607 \\
LF+MF $\to$ HF, $N=1$ & 5.6988 & 5.1754 \\
LF $\to$ HF, $N=10$ & 6.0100 & 5.7142 \\
MF $\to$ HF, $N=10$ & 6.4182 & 5.8559 \\
LF+MF $\to$ HF, $N=10$ & 5.9208 & 5.3785 \\
\bottomrule
\end{tabular}
\end{table}